\documentclass[10pt]{article}

\usepackage{amsmath,amsthm,amsfonts,mathrsfs,anysize,fancyhdr,epsfig,mdframed, float,graphicx,dsfont}

\usepackage{chngcntr}
\usepackage{algorithm,algorithmic}
\theoremstyle{plain}
\newtheorem{theorem}{Theorem}

\usepackage{enumitem}
\theoremstyle{definition}
\newtheorem{definition}[theorem]{Definition}

\newtheorem{remark}[theorem]{Remark}

\usepackage{chngcntr}

\usepackage[paperwidth=8.5in,paperheight=11in,top=1.2in, bottom=1.2in, left=0.9in, right=0.9in]{geometry}
\usepackage{mathtools}                                                      % need for `show only references'
\mathtoolsset{showonlyrefs=true}                                    % only equations which are labeled AND referenced will be numbered.
\begin{document}
\title{Generalization in fully-connected neural networks for time series forecasting}
\author{Anastasia Borovykh\thanks{CWI Amsterdam, the Netherlands. \textbf{e-mail}: anastasia.borovykh@cwi.nl}, \; Cornelis W. Oosterlee \thanks{CWI Amsterdam, the Netherlands. \textbf{e-mail}: c.w.oosterlee@cwi.nl}, \; Sander M. Boht\'e \thanks{CWI Amsterdam, the Netherlands. \textbf{e-mail}: s.m.bohte@cwi.nl}
}
\maketitle

\begin{abstract}
In this paper we study the generalization capabilities of fully-connected neural networks trained in the context of time series forecasting. Time series do not satisfy the typical assumption in statistical learning theory of the data being i.i.d. samples from some data-generating distribution. We use the input and weight Hessians, that is the smoothness of the learned function with respect to the input and the width of the minimum in weight space, to quantify a network's ability to generalize to unseen data. While such generalization metrics have been studied extensively in the i.i.d. setting of for example image recognition, here we empirically validate their use in the task of time series forecasting. Furthermore we discuss how one can control the generalization capability of the network by means of the training process using the learning rate, batch size and the number of training iterations as controls. Using these hyperparameters one can efficiently control the complexity of the output function without imposing explicit constraints. 
\end{abstract}

\section{Introduction}
Forecasting time series is an exceptionally difficult task due to the risk of overfitting on the dataset, in particular in the case of overparametrized networks \cite{zhang16}, \cite{zhang98}. In other words, when using the past to predict the future one has to be certain to have succeeded in extracting a signal from the past that will propagate to the future, and not simply fitted a complex function on the past. Neural networks, while being powerful function approximators that are relatively easy to optimize, can lead to poor extrapolation in time series forecasting due to the latter. Due to their ability to approximate almost any function it is of the essence to ensure that the network is learning the signal of interest instead of the noise. Understanding the structure of neural networks and the ability of a trained network to perform well on unseen data is therefore of utmost importance, and the main objective of this paper. 

The loss surface of a neural network, defined as a function of the loss over the weights, is typically highly non-convex and can, for a deep network, depend on a large number of parameters (the weights). Even for a simple network, the number of local minima and saddle points in the loss surface may grow exponentially in the number of parameters. The general shape of this loss function, and also the differences in the loss functions of small and large neural neural networks, is an active topic of research  \cite{choromanska15}, \cite{li17}. In terms of theory, a recent line of work has related the neural network loss surface to Gaussian random fields \cite{choromanska15}, \cite{auffinger13}, \cite{bray07}, \cite{dauphin14}. Alternatively, random matrix theory has been used to obtain insight into the loss surface \cite{pennington17}. In more empirical lines of work, the authors of \cite{li17} found that adding more layers to a network gives rise to a more non-convex loss surface, so that adding more layers can complicate the training of the neural network by causing the optimisation methods to get stuck in sub-optimal critical points. 

The above work gives insight into the structure of the loss surface on the training dataset. For noisy time series, a trained network is able to generalize well, that is perform well on unseen data, when it is not overfitting on noise in the training dataset. However, as is mentioned by the authors of \cite{zhang16}, if the network is big enough (i.e. overparametrized) it can even fit a random noise dataset almost perfectly, but it will most certainly have bad performance out of sample. Understanding the structure of the minima so that a network will perform well on unseen data can give insight into setting up the training methods, for example to avoid convergence on noise. %It was shown that the width of the minima is a good indicator of its ability to generalize \cite{hochreiter97}, \cite{keskar16}, \cite{wu17}. Alternatively, the smoothness of the learned function, as measured by e.g. the Jacobian with respect to the input can also be a good indicator \cite{novak18}. Using these properties one can bias the optimisation algorithm to end up in better minima, as is done in the work of e.g. \cite{choromanska17}, \cite{kenton17}. 

There are different ways of measuring the learning capability of a neural network (see \cite{bernier00} for an overview). One is the output sensitivity, or the first derivatives i.e. the Jacobian, of the error with respect to the input (see \cite{novak18}) or the weights. The other measure is the statistical sensitivity which evaluates the output range variation of a node when its inputs or weights are perturbed. As is shown in \cite{bernier00}, this is equivalent to considering second derivatives, the Hessian, of the loss function with respect to the input or weight values. The statistical sensitivity with respect to the weights gives a measure for the smoothness of the error surface, with a small value of the statistical sensitivity implying a small output variation when weights are perturbed. The sensitivity with respect to the inputs measures the input noise immunity. Using the statistical sensitivity as a measure for generalization is intuitive in the sense that we are interested in the robustness of the network when the input is perturbed. It has also been proposed that the Hessian with respect to the weights can be used as a measure for generalization. These flat minima in the weight space correspond to simpler functions learned \cite{hochreiter97} or can be related to the Bayesian evidence \cite{smith18}. 

Regularisation in the network can help to obtain a learned function with lower complexity such that a better generalization may be obtained. Typical explicit regularisation methods, such as $L_1$ or $L_2$ regularisation or multiplicative noise injection (such as dropout \cite{srivastava14}), contribute to the generalizability of the trained function by restricting the function complexity in some way. %Weight decay ($L_2$ regularization) however assumes that the optimal parameters are close to zero, therefore being restrictive since it does not consider solutions away from zero with better test performance. 
For the regularisation method to work well, we need to understand how to make a trade-off between the complexity of the function and its ability to fit the data. This trade-off is known as the information bottleneck \cite{tishby15}, and we study this in the coming sections to understand the effects of the trade-off on the learned function. Alternatively to explicit regularization, the noise in a stochastic gradient descent method (SGD) can act as an implicit regularizer. The gradient is computed over batches and, as opposed to computing the full gradient, SGD thus introduces non-isotropic noise into the optimisation scheme. It can be shown that this drives the parameters away from sharp minima towards the broader ones. In particular, the noise variance is proportional to the learning rate over the batch size, so that a large learning rate and small batch size result in a higher noise component. This has been shown in previous work, e.g. \cite{smith18}, \cite{chaudhari18} and will be a focus of this work as well.

The novelty of our contribution consists in a thorough analysis of what generalizability means for time series forecasting with fully-connected neural networks. In particular, time series do not satisfy the typical assumption in statistical learning theory of i.i.d. data. Furthermore, while generalizibility for image datasets has been studied extensively, the problem is much more complex for time series: the dataset is typically much smaller, the signal-to-noise ratio might be low, the distribution can be non-stationary and there is little intuitive indication of what the underlying pattern in the data must be. Due to the non-i.i.d. nature the pattern might also change through time. Understanding what it means for a neural network to have good generalizibility, i.e. learning a consistently present pattern instead of overfitting on noise or on a changing pattern, and how this can be achieved through the learning algorithms will be the main task of this paper. 

We assume that the reader is familiar with the general neural network concepts such as optimisation methods like stochastic gradient descent and its parameters and the neural network architectures. For a general introduction to this we refer to \cite{bishop07}. The rest of this paper is structured as follows: in Section \ref{sec2} the loss surface structure is studied in a simplified setting; in Section \ref{sec3} the input and weight Hessians are introduced as generalization metrics and the relation between the two is described; in Section \ref{sec5} it is discussed how to make the trade-off between complexity and data fit and how one can influence complexity during the training of the network; finally Section \ref{sec6} presents the numerical results. 

%\blu{Verschil L1 en L2 norm voor gladheid?????????????????}

\section{Loss surface structure}\label{sec2}
In this section we give some background about neural networks and the properties of their loss surfaces and the implications of this structure for generalization capabilities.  
\subsection{Loss surface as a Gaussian random field}\label{sec21}
The loss surface of a neural network is defined as the loss function over the high-dimensional weight space. This loss surface can be related to a Gaussian process on a high-dimensional space \cite{dauphin14}, \cite{choromanska15}. With this insight, one is able to obtain theoretical results on the structure of the loss surface. We shortly repeat this derivation and discuss its implications. 
Let the inputs to the  neural network be given by $x\in\mathbb{R}^{n_0}$. Let $W^{(l)}\in\mathbb{R}^{n_{l}\times n_{l-1}}$ be the weight matrix in layer $l$ with element $w_{i,j}^{(l)}$ connecting neuron $i$ in layer $l$ and $j$ in layer $l-1$. Define $w$ as the vectorized total weights in the network, so that $w\in\mathbb{R}^{d}$ with $d=n_0 n_1 +n_1 n_2 +... +n_{L-1} n_L$, with $d$ thus being the dimension of the weight space. In the rest of this paper the dimension $\mathbb{R}^d$ refers to \emph{column} vectors. The first layer output, for $l=1$, is given by
\begin{align}
a^{(1)} = f(z^{(1)}) = f\left(W^{(1)}x\right),
\end{align}
where $f(\cdot)$ is the non-linear activation function, $a^{(1)}\in\mathbb{R}^{n_1}$ is the activation of the first layer and $z^{(1)}\in\mathbb{R}^{n_1}$ is the pre-activation output. 
Each subsequent layer $l=2,...,L$ outputs, 
\begin{align}
a^{(l)} = f(z^{(l)}) = f\left(W^{(l)}a^{(l-1)}\right).
\end{align}
The final layer output is then given by, 
\begin{align}\label{eq:output1}
\hat y(x,w)=qf(W^{(L)}f(W^{(L-1)}...f(W^{(1)}x)))...),
\end{align}
with $q$ a scaling factor. Assume the data is given as a set of inputs $x$ and outputs $y$ generated from some data-generating distribution $\mathcal{D}$. Typical loss functions are the mean absolute error,
\begin{align}\label{eq:lossfn1}
\mathcal{L}(x, w,y) = \mathbb{E}_{(x,y)\sim\mathcal{D}}[|\hat y(x,w)-y|],
\end{align} 
or the mean squared error,
\begin{align}\label{eq:lossfn2}
\mathcal{L}(x, w,y) = \mathbb{E}_{(x,y)\sim\mathcal{D}}[(\hat y(x,w)-y)^2].
\end{align} 
where the expected values are taken over the data generating distribution.

We define a critical point $w^*$ and its index $\alpha$ as follows,
\begin{definition}[A critical point and its index]\label{def1} A critical point $w^*$ of some differentiable function $g:\mathbb{R}^d\rightarrow\mathbb{R}$ is point $w^*\in \mathbb{R}^d$ where all partial derivatives of the function $g$ are zero. In this work, we also refer to a critical point in a more loose definition as the point to which the optimization algorithm for the neural network has converged. For a function of $d$ variables, the number of negative eigenvalues of the Hessian matrix $H^w$, the matrix of second-order derivatives of the loss function with respect to the parameters (defined more explicitly in \eqref{eq:hesspar}), at a critical point $w^*$ is called the index $\alpha$ of the critical point.
\end{definition} 

Following the derivation in \cite{choromanska15}, let the non-linear activation function be the rectified linear unit defined as $f(x)=\max(x,0)$ and replace the activation function in \eqref{eq:output1} by the term $A_{i,j}\in\{0,1\}$, which denotes whether a path $(i,j)$, where $j$ labels any of the $P$ paths from the input $i$ to the output, is active or not. We obtain,
\begin{align*}
\hat y(x,w) = q\sum_{i=1}^{n_0}\sum_{j=1}^Px_iA_{i,j}\prod_{k=1}^L w_{i,j}^{(k)}.
\end{align*}
Here $x_i$ refers to the $i$-th element of the input vector $x$ and $P:=n_1n_2\dots n_L$ is the number of paths from a given network input to its output.

We now make the first key assumption that each path is equally likely to be active. The probability of a path being active follows a Bernoulli distribution with probability $\rho$, independent of the input. Taking the expected value over the activation we obtain,
\begin{align}\label{eq:output2a}
\mathbb{E}_A[\hat y(x,w)] = q\sum_{i=1}^{n_0}x_i\rho\sum_{j=1}^{P}\prod_{k=1}^Lw_{i,j}^{(k)},
\end{align}
with $i$ the summation over the inputs and $P=n_1\dots n_L$ representing the summation over the further possible paths in the network. We remark that this expression is similar to a deep \emph{linear} model multiplied by the factor $\rho$. 

The second key assumption in this section is to let the input elements be sampled independently as $x_i \sim\mathcal{N}(0,\sigma_I^2)$ (and let $\sigma_I^2=1$ for simplicity). Due to the summation being over independent standard Gaussian random variables, $\hat y$ is equal to a Gaussian process on the weight space. %Assuming that there is redudancy in the network allows to skip superscript $(k)$ in the weights and assuming uniformity ensures that all ordered products of unique weights appear the same number of times. We obtain, 
%\begin{align}\label{eq:output3a}
%\mathbb{E}[y(w,x)|x] \approx q\sum_{i_1,...,i_L=1}^{\Lambda}x_{i1,...,i_L}\rho\prod_{k=1}^Lw_{i_k},
%\end{align}
%where $\Lambda=\sqrt[L]{P}$ with $P$ being the total number of all paths between the network input and output and $x_{i1,...,i_L}$ are independent, identical and standard normally distributed. Define $q:=\Lambda^{-(L-2)/2}$ and assume the weights satisfy a spherical constraint $\frac{1}{\Lambda}\sum_{n=1}^\Lambda w_n^2 = 1$.
Letting the loss function be given by the absolute loss as in \eqref{eq:lossfn1} in which the expected value can be taken over the activations, %\blu{blijft vaag waarom oppeens die activations; die hangen af van de data}
\begin{align}
\mathcal{L}(x, w,y) = \mathbb{E}_A\left[|\hat y(x,w)-y|\right],
\end{align} 
due to the $x_i$ being sampled from a $\mathcal{N}(0,1)$ distribution, this loss function follows a Gaussian process distribution. For a particular value of $q$ it is equal to the well-studied Hamiltonian of spin-glass systems \cite{auffinger13} and previous work on Gaussian random fields can be applied \cite{bray07}, \cite{fyodorov07} to gain insight into the structure of the critical points. 

\subsection{Structure of the critical points}
In this section we briefly summarize the results on the loss surface structure of Gaussian random fields in high dimensions. The works of \cite{bray07} and \cite{fyodorov07} show that for Gaussian random fields on high-dimensional spaces the critical points of the surface posess a particular structure. In \cite{bray07} the authors show, by means of a generalized Kac-Rice formula, a linear dependence between the index of a critical point and its loss value (the error $\mathcal{E}$). A similar result can be obtained for neural networks as is done in \cite{auffinger13}, \cite{choromanska15}. Let $\Lambda$ be the number of different weights in the network, which is assumed to be the $L$-th root of the total number of paths from input to output in the network,
\begin{align}
\Lambda :=\sqrt[L]{n_0P}.
\end{align} 
Under the assumptions made in Section \ref{sec21}, the loss surface of a neural network on a \emph{high-dimensional} parameter space, in other words for deep and wide networks or as $\Lambda$ increases, has the following properties,
\begin{enumerate}[label={(\arabic*)}]
\item let $\mathcal{E}_0<\mathcal{E}_{m_1}<\mathcal{E}_{m_2}<...$; there exists a layered structure of critical points: critical values in a band $(\mathcal{E}_0,\mathcal{E}_{m_1})$ above the global minimum $\mathcal{E}_0$ are more likely to be local minima, the band $(\mathcal{E}_{m_1},\mathcal{E}_{m_2})$ consists of local minima and saddle points of index 1, the band $(\mathcal{E}_{m_2},\mathcal{E}_{m_3})$ consists of local minima and saddle points of index 1 and 2, and so on;
\item local minima dominate over saddle points in a band of values close to the global minimum;
\item high-index critical points lie at high loss levels; in other words, a high value of $\alpha$ corresponds to a high loss level $\mathcal{E}$.
\end{enumerate}
To conclude, by making several assumptions on the activation function and the distribution of the inputs, it is possible to relate the neural network loss function to a particular kind of Gaussian random field, as commonly encountered in spin-glass systems. By an application of the Kac-Rice theorem, one is able to obtain a relationship between the index of a critical point of this Gaussian random field and its value. It can be shown that high-index saddle points lie at high loss levels, while local minima are close to the global minimum. %If the gradient descent algorithm used to optimize the neural network weights is very suitable for escaping from high-index saddle points, i.e. those with many directions facing downwards, then the optimal weights found by the optimization algorithm will have a loss value close to that of the global minimum. This could explain why neural networks tend to be relatively `easy to optimize'. 

\subsection{Loss and the entropy}
As was shown in the previous section, under certain -- albeit restrictive -- assumptions on the deep neural network, the loss surface is given by a Gaussian random field on a high-dimensional space; this space represents the weight space and its dimension is given by the number of weights in the network as determined via the number of nodes per layer and the number of layers used. Gaussian random fields on high-dimensional spaces posess a particular structure of the locations of the critical points in the asymptotic setting. Similarly, there exists a result on the entropy of these critical points. We recall here a result on the width of the minima, as stated in \cite{becker18}. To measure the width of a minimum $w^*$, consider the entropy which is defined as,
\begin{align}
S(w^*) = -\log\det (H^w(\mathcal{L}(x,w^*,y)),
\end{align}
with $H^w$ being the Hessian matrix. A larger entropy means larger basin volume, or a wider minimum. We then state the following Theorem on the width of the minima, as is given and proved in \cite{becker18}.

\begin{theorem}[Expected entropy \cite{becker18}]\label{thm2}
Let $\mathcal{E}$ be some loss level. The expected entropy of the Hessian of the loss function that takes value $\Lambda \mathcal{E}$ asymptotically, has the following expected entropy %\blu{have to explain all the parameters}
\begin{align}
\mathbb{E}\left[S(w^*)|\Lambda\mathcal{E}\right] \simeq & -(\Lambda-1)\log(\rho) + \frac{\Lambda-1}{2}\log\left(\frac{\Lambda}{2(\Lambda-1)L(L-1)}\right)\\
&-\frac{\Lambda-1}{\pi}\int_{-\sqrt{2}}^{\sqrt{2}}\log \bigg|\sigma \sqrt{\frac{\Lambda}{\Lambda-1}}\frac{\mathcal{E}}{\rho}-t\bigg|\sqrt{2-t^2}dt,
\end{align}
where $L$ is the number of layers and $\rho$ is the probability of the Bernoulli distributed weights being one. 
\end{theorem}
The above theorem gives a relation  between the number of layers in the network, the loss level at a particular point in the weight space and the width at that point in the weight space. In particular, the lower the train loss the lower the entropy and thus the sharper the minima. In other words, wider minima lie at higher loss values in the loss surface. This seems intuitive: in order to obtain a low train loss, one has to fit a more complex function which passes through all the obervations. A good fit to the training data is however not sufficient to obtain good out of sample performance; we will discuss the concept of generalization in the next section. %Furthermore from this expression it follows the trade-off between loss and width is less severe for deeper networks. In other words, as the number of layers increases, the minima that have a low loss value have approximately the same width as minima with a higher loss value. \blu{Zal ik hier zelf nog een plaatje toevoegen zoals fig4 in becker?}
%\blu{is dit te gebruiken om de learning rate te bepalen zodat je nog goede generalization kan hebben of is deze formule te generiek? Je wilt eigenlijk de laagste loss hebben waarmee je nog goede generalizatie kan hebben? dus wat is dan een hessian waarmee we nog goede generalizatie hebben? oftewel is er een relatie tussen de hessian en loss op train set en de loss op test set? dit gaat sws niet data onafhankelijk zijn.. } 

\subsection{Generalization}
%\subsubsection{Functions with noise}
Since the data generating distribution $\mathcal{D}$ is typically unknown, in extrapolation problems one assumes to have access to samples $(x^i,y^i)$ drawn (i.i.d.) from this distribution. One assumes $y^i = g(x^i)+\epsilon^i$, for some unknown function $g(\cdot)$; so that the $y^i$ are noisy observations of the true function of interest. In our setting we are interested in time series forecasting, i.e. we have $x^i=(y^{i-n},...,y^{i-1})$ where $i$ is the time index, so that $n$ historical points of $y$ are used to predict its future, in this setting one-day-ahead, value. This can be extended to $x$ containing the historical observations of multiple time series used to forecast $y$. Note that here the i.i.d. assumption is violated since the observations should clearly be dependent. Nevertheless, using these datapoints as input a neural network can be used to extract a meaningful repeating pattern from the dataset. Note that the $y^i$'s, and thus the $x^i$'s, are \emph{noisy} observations. We define the sample loss function as the loss function on that dataset, i.e. for the squared loss we obtain,
\begin{align}\label{eq:lossfn}
\mathcal{\hat L}(x, w,y) = \frac{1}{N}\sum_{i=1}^N\left(\hat y(x^i,w)-y^i\right)^2,
\end{align}
where $(x^i,y^i)$ for $i=1,...,N$ is the train dataset and $\hat y(x^i,w)$ the neural network output. 

generalization is the relationship between a trained networks' performance on train data versus its performance on test data. This is a highly desirable property for neural networks, where ideally the performance on the train data should be similar to the performance on similar but unseen test data. In general, the generalization error of a neural network model $\hat y(x,w)$ can be defined as the failure of the hypothesis $\hat y(x,w)$ to explain the dataset sample. It is measured by the discrepancy between the true error and the error on the sample dataset, 
\begin{align}\label{eq:genbound}
\mathcal{L}(x,w,y) - \mathcal{\hat L}(x,w,y).
\end{align}
In statistical learning theory a bound on this error is typically dependent on the complexity of the hypothesis class where the hypothesis $\hat y(x,w)$ is in, as well as on the number of samples in the dataset. Obtaining bounds on this error is a topic of active research with recent advancements including \cite{dziugaite17}, \cite{zhou18} where the authors use PAC-Bayes theory. In the rest of the paper we will use the notation $\mathcal{L}(x,w,y)$ to denote the empirical loss function as computed on the sampled data. 

Typically, a trained network is able to generalize well when it is not overfitting on noise in the train dataset. Since neural networks are known to be universal approximators and thus -- when the network is large enough -- are able to approximate any function, when training one aims to extract a meaningful pattern in the data instead of learning a flexible function that is able to fit all training points. In particular in the setting of overparametrized networks it is easy for the network to fit the training points, however being able to avoid this overfitting is an essential task. A somewhat straightforward way to define the generalization capability is to study the robustness of the network with respect to input perturbations. For some input perturbation $\epsilon\in\mathbb{R}^{n_0}$, the change in the loss function should be small,
\begin{align}
|\mathcal{L}(x+\epsilon,w,y) - \mathcal{L}(x,w,y) |<\delta.
\end{align}
When the neural network is heavily overfitting on the noise, a small change in the input parameters might result in a large change in the neural network output. In this setting the generalization is related to a smoothness assumption on the function output. In the coming sections our goal is to understand and be able to control the generalization of neural networks in the overparametrized, deep neural networks for time series forecasting, a setting in which the signal in the series can be weak and we lack the availability of large datasets. We aim to define metrics that can be used to measure when a learning algorithm can be expected to perform well. 

%\subsubsection{Deterministic functions}
%In an alternative setting, a neural network can be used to learn some complex deterministic function. This is known as \emph{interpolation}. This is for example the case when neural networks are used as methods for solving partial differential equations \cite{sirignano17}, \cite{jentzen17}, \cite{raissi17}, where the output is e.g. some function $u$ of e.g. time and space. In this case the neural network is trained on noise-free data, i.e. the $y_i$ are noiseless obervations of the function of interest $u$. In this situation the danger of overfitting lies in finding a high-complexity function as in the left-hand plot of Figure \ref{fig0}, and the generalizability is quantified again by some smoothness assumption on the function output. While finding the global minimum on the train data is not the objective in the noisy obervation case, when fitting a deterministic function on noisefree data the global minimum is what is aimed for, and the optimization algorithm should be tuned to finding this minimum, while still satisfying the function smoothness requirements. In this work we will mostly focus on the functions with noise, however understanding and finding the minima that generalize well in the case of deterministic functions will be of interest in future work. 

\section{Metrics for measuring the generalization}\label{sec3}
In this section we present metrics, the input and weight Hessians, for measuring the generalization capabilities. 
\subsection{The weight Hessian}
The Hessian with respect to the weights will be used in order to obtain insight on the noise robustness of the weights, giving a metric for measuring the networks' capability to generalize well to unseen data. 
\subsubsection{Definition}
The Hessian $H^w(\mathcal{L})\in\mathbb{R}^{d\times d}$ of the loss function with respect to the weights has elements 
\begin{align}\label{eq:hesspar}
h_{ij}^w=\partial_{w_i}\partial_{w_j}\mathcal{L}(x,w,y),
\end{align}
which represents the rate of change of the derivative with respect to $w_j$ in the direction of $w_i$. The Hessian thus represents the curvature of the loss surface of the neural network. The eigenvectors and eigenvalues represent the direction and curvature in that direction, respectively. For the large neural networks typically used in image processing, computing and storing the Hessian can be very time-consuming. In the case of time series forecasting the networks used will be smaller, but nevertheless the Hessian can contain thousands of elements. In the rest of this paper we will sometimes drop the dependence of the loss function on the input $x$ in case we are interested in the effects of the weights only. 

The Hessian gives insight into the flatness of the minimum, and, as we will show in Section \ref{sec42}, this can be related to input noise resistance of the output function. In this sense the Hessian can be related to the minimum description length, where a Hessian with small eigenvalues corresponds to a simpler function being learned. Alternatively, the Hessian is used in second-order optimization methods where the step size in each direction is inversely proportional to the curvature in that direction: in directions with large curvature it takes small steps, while in directions with small curvature it takes larger steps \cite{martens10}. %This is also the intuition behind methods such as momentum. 
\subsubsection{Learning rate, batch size and the Hessian}\label{sec312}
It has been mentioned in prior research \cite{kenton17}, \cite{mandt17}, \cite{smith18}, \cite{chaudhari18} that a relationship exists between the test error and the learning rate and batch size used in the SGD updating scheme. In this section we obtain a similar conclusion through a slightly different derivation. Let the gradient in a mini-batch $\mathcal{S}$ be $g_\mathcal{S}\in\mathbb{R}^d$ and the full gradient be $g\in\mathbb{R}^d$, where $d$ is the weight space dimension, defined respectively as,
\begin{align}
g_{\mathcal{S}} := \frac{1}{M}\sum_{i\in\mathcal{S}}\nabla_w\mathcal{L}(x^i,w,y)=:\frac{1}{M}\sum_{i\in\mathcal{S}}g_i,\;\;\; g:=  \mathbb{E}_{(x,y)\sim\mathcal{D}}\left[\nabla_w\mathcal{L}(x,w,y)\right].
\end{align}
The weight update rule is given by,
\begin{align}
w_{t+1} = w_t - \eta g_{\mathcal{S}},
\end{align}
where $\eta$ is the learning rate. By the central limit theorem, if $(x_i,y_i)\sim\mathcal{D}$ i.i.d. then,
\begin{align}
(g_{\mathcal{S}}-g)\sim\mathcal{N}\left(0,\frac{1}{M}K\right),\;\; K=\mathbb{E}\left[( g_{i}-g)^T(g_{i}-g) \right]\approx \frac{1}{N-1}\sum_{i=1}^N(g_i-g)^T(g_i-g).
%=\frac{1}{N-1}\sum_{i=1}^Ng_i^Tg_i-g^Tg.
\end{align}
Note that the approximation of the noise by a Gaussian distribution holds in the limit of the sample size tending to infinity and when the gradients for the batches are not heavy-tailed. Although the sample size is typically finite and the gradient distribution can be heavy-tailed, the approximation is widely used.

The weight update rule can then be rewritten as,
\begin{align}\label{eq:wupnoise}
w_{t+1} = w_t - \eta g - \frac{\eta}{\sqrt{M}} \epsilon,
\end{align}
where $\epsilon\sim\mathcal{N}(0,K)$ with $\epsilon\in\mathbb{R}^d$. If convergence has been reached,
\begin{align}\label{eq:ass1}
|\mathcal{L}(w_{t+1}) - \mathcal{L}(w_t)|<\delta.
\end{align}
Note that for ease of notation we have omitted the dependence of the loss function on the input data $(x,y)$.
By a Taylor expansion method for the loss function evaluated on the full training data we have,
\begin{align}\label{eq:tayexp}
\mathcal{L}(w_{t+1}) \approx \mathcal{L}(w_t) -\left(\eta g - \frac{\eta}{\sqrt{M}} \epsilon\right)^T  \nabla_w \mathcal{L}(w) + \frac{1}{2}\left(\eta g - \frac{\eta}{\sqrt{M}} \epsilon\right) ^T H^w(\mathcal{L}(w_t)) \left(\eta g - \frac{\eta}{\sqrt{M}} \epsilon\right),
\end{align}
where $\nabla_w \mathcal{L}(w)\in\mathbb{R}^d$ denotes the gradient of the total loss and $H^w(\mathcal{L}(w))\in\mathbb{R}^{d\times d}$ the Hessian of the total loss. Note that we thus approximate the loss surface by a quadratic function.
Then, using the convergence in \eqref{eq:ass1} and the Taylor expansion for the loss function in \eqref{eq:tayexp}, we obtain, %\blu{dubbelcheck}
\begin{align}
H^w(\mathcal{L}(w_t))\approx \frac{\delta + 2\nabla_w\mathcal{L}(w_t)^T \left(\eta g - \frac{\eta}{\sqrt{M}} \epsilon\right)}{\big|\eta g - \frac{\eta}{\sqrt{M}} \epsilon\big|^2} = \frac{\delta + 2\nabla_w\mathcal{L}(w_t)^T \left(g - \frac{1}{\sqrt{M}} \epsilon\right)}{\eta \big| g - \frac{1}{\sqrt{M}} \epsilon\big|^2}.
\end{align}
From this expression we see that the Hessian at convergence is small if a large learning rate or a small batch size has been used. %This result has also been mentioned in previous research in which it was mentioned to use an as large learning rate as possible with which convergence can still be obtained. 
Our simple derivation shows that if convergence is obtained, the Hessian in that point in weight space is smaller for larger learning rates and smaller batch sizes, i.e. the Hessian is inversely proportional to the fraction $\eta/\sqrt{M}$. In case of using full batch gradient descent, thus if $\epsilon=0$, the relationship between convergence and the learning rate remains the same: convergence achieved with a large learning rate results in a smaller Hessian which in turn corresponds to a wider minimum. %\blu{is er een soort optimale learning rate? i.e. kunnen we uit de loss surface structuur afleiden wat voor learning rate goed zal zijn om te gebruiken? }

%\subsubsection{Sample size and Hessian}
%Consider the second order derivative of the loss function $l(f(w))$,
%\begin{align}
%\nabla^2 l(f(w)) = l''(y(w))\nabla y(w)\nabla y(w)^T + l'(y(w))\nabla^2 y(w).
%\end{align}
%Note that $l'(y(\cdot))$ denotes the derivative with respect to its input parameter $y(\cdot)$. Then, due to the convexity of the loss function $l''(\cdot)>0$ and the Hessian of the loss function as given in \eqref{eq:lossfn} can be written as,
%\begin{align}\label{eq:hessloss}
%\nabla^2\mathcal{L}(w) = \frac{1}{N}\sum_{i=1}^N (\sqrt{l''(y(w))}\nabla y(w))(\sqrt{l''(y(w))}\nabla y(w))^T + \frac{1}{N}\sum_{i=1}^N l'(y(w))\nabla^2 y(w).
%\end{align} 
%If we would take for $l(y(w)$ the squared loss, i.e. $l(y(w)) = (y(w)-y_i)^2$, then $f'(y(w))=2(y(w)-y_i)$ and $f''(y(w)) = 2$.
% 
%We are interested in obtaining results on the spectrum of eigenvalues of the Hessian, since this amount of positibe outliers was shown to be a metric for the generalizeability of the weight configuration. The right-hand side of the expression in \eqref{eq:hessloss} is the sample covariance matrix of the scaled gradients. Recent work of \cite{bloemendal16} related the spectrum of this sample covariance matrix to that of the population matrix. \blu{Hessian wordt kleiner als meer samples? Assume dat de gradient overal net anders is en netto nul oid? Zitten er dan nog restrictions vast aan de data? e.g. correlated met zo veel covariance?} 

\subsubsection{The weight Hessian and generalization}\label{sec23}
Generalization in our setting refers to a kind of robustness of the trained network. More specifically, particular transformations of the input do not decrease the accuracy of the classification/forecast as computed on the train set. In other words, when a network has been trained on a set of patterns, certain transformations of these patterns should still be interpreted correctly. A higher robustness to input noise, which can be measured by Jacobians/Hessians with respect to input or weights, leads to better generalization, i.e. the smaller the Hessian the wider the minimum. 

Consider the eigenspectrum of the Hessian, i.e. the set of eigenvalues $(\mu^i)$, $i=1,...,d$, determined via,
\begin{align}
H^w v = \mu v,
\end{align}
where $v$ is the eigenvector corresponding to the eigenvalue $\mu$. If the eigenvalues of $H^w$ are positive (resp. negative) at some critical point $w^*$, that point is a local minimum (resp. maximum), and if the critical point has both positive and negative eigenvalues it is called a saddle point (see also Definition \ref{def1} on the index of a critical point). The eigenvector corresponding to the largest eigenvalue of $H^w$ indicates the direction of greatest curvature of the loss function. The size of the positive eigenvalues is thus a measure for how well a minimum will generalize to unseen data. A large positive eigenvalue in the direction of the corresponding eigenvector thus means that a sharp increase in loss will occur in that direction in weight space. If a minimum is wide, and thus has small eigenvalues in many directions, the minimum is better resistant to noisy transformations of the weights, while a sharp minimum has a higher sensitivity to the noise in the weights. A sharp minimum is thus said to have overfitted on the noise in the training dataset, while a wider minimum may imply that a `simpler' and more robust function has been learned. 

In the work of \cite{hochreiter97} the authors show that flat minima correspond to a minimization of the expected description length of the neural network function induced by the weights. The authors of \cite{smith18} show a relationship between the Bayesian evidence and the Hessian, showing that maximizing the evidence corresponds to a minimization of the Hessian, by approximating the evidence with a Taylor expansion of the cost function. 

%\subsubsection{generalization from a Bayesian perspective}
%\cite{smith18} shows that Bayesian evidence is related to Hessian; repeat here for completeness of een MDL argument zoals \cite{hochreiter97}.\blu{TO DO; note this derivation will probably only hold for i.i.d. so mention that!!! extension to non-iid is not what we do here}
\subsubsection{Downsides of the weight Hessian}\label{sec314}
A metric that is commonly used to measure the width of the minimum is the trace. The trace of a squared matrix $H$ of size $d\times d$ is defined as,
\begin{align}
Tr(H) :=\sum_{i=1}^d h_{ii} = \sum_{i=1}^d \mu_i,
\end{align}
where $h_{ii}$ denotes the elements on the diagonal of the Hessian and $\mu_i$ denotes the eigenvalues. 
While prior research has noted that a lower Hessian (in terms of trace or some other norm) leads to better generalization, there has also been contradicting evidence. One critique on using the trace of the weight Hessian for measuring generalization capabilities is that the Hessian can be scaled in such a way that the output function remains the same but the trace of the Hessian can become large or small. This is the conclusion of the work of \cite{dinh17}. Consider a neural network with two layers, so that the output is given by
\begin{align}
\hat y(x,w) =f( W^{(1)}x)W^{(2)}.
\end{align}
Note that if $f(\cdot)$ is the rectified linear unit $f(x) = \max(x,0)$, then we can scale the weights by a constant $\alpha>0$ without changing the output as follows,
\begin{align}\label{eq:scaling}
\hat y(x,w) =f( \alpha W^{(1)} x) \alpha^{-1} W^{(2)}.
\end{align}
The gradient and Hessian of the loss $\mathcal{L}$ with respect to the weights $w$ can be modified by $\alpha$. We have,
\begin{align}
\mathcal{L}(W^{(1)},W^{(2)}) = \mathcal{L}(\alpha W^{(1)},\alpha^{-1} W^{(2)}).
\end{align}
Then
\begin{align}
\nabla_w \mathcal{L}(\alpha W^{(1)},\alpha^{-1} W^{(2)}) = \nabla_w \mathcal{L}(W^{(1)},W^{(2)})\begin{bmatrix}\alpha^{-1} & 0\\ 0&\alpha \end{bmatrix},
\end{align}
where the derivatives are taken with respect to $\alpha W^{(1)}$ and $\alpha^{-1}W^{(2)}$.
Similarly,
\begin{align}
H^w( \mathcal{L}(\alpha W^{(1)},\alpha^{-1} W^{(2)})) = \begin{bmatrix}\alpha^{-1} & 0\\ 0&\alpha \end{bmatrix}H^w(\mathcal{L}(W^{(1)},W^{(2)}))\begin{bmatrix}\alpha^{-1} & 0\\ 0&\alpha \end{bmatrix}.
\end{align}
For large weights, e.g. when $\alpha$ is large, the second derivative with respect to these weights is smaller, however if the next-layer weights are scaled by $\alpha^{-1}$ the output function has not changed. Therefore, due to the many symmetries existing in a neural network, there exist symmetries that will not modify the output function and the generalization capabilities, but are able to scale the Hessian with respect to particular weights to be larger or smaller. Norms like the trace norm or the Frobenius norm are thus not enough to determine the generalization capabilities \emph{if} one compares minima that are equivalent through these symmetries. % Similarly in e.g. \cite{kenton18} a sharp minimum as determined by a large Frobenius norm of the Hessian had similar generalization capability as a wide minimum. %In this work the authors showed that by taking smaller steps in the directions of sharpest descent one obtains sharper minima but similar generalization performance. This may be explained by the fact that not only the norm of the full Hessian should be small, but in particular in the directions tangent to the directions of the input transforms the eigenvalues should be small, or the loss function in those directions should remain low. 

This analysis thus shows that for each individual minimum found, a large trace of the Hessian is not informative. In \cite{kenton17} the authors claim that while these sharp minima with a generalization similar to a wide minima exist, SGD does not converge to these minima and for the minima found by SGD the width correlates well with generalization. We obtain a similar result in the numerical experiments in Section \ref{sec6}. This can be explained by the fact that the Hessians' sensitivity to scaling becomes an issue for $\alpha<<1$ or $\alpha>>1$, i.e. when the scaled and unscaled minima are far away in weight space, because then a minimum with a large Hessian norm can have similar generalization as a minimum with a small norm. %In this case the weight norm of $(W^{(1)}, W^{(2)})$ will differ significantly from $(\alpha W^{(1)}, \alpha^{-1}W^{(2)})$. So for similar weight norms the Hessian should remain a valid metric of generalizability. 
We claim that for an SGD algorithm started from random initializations, if these initializations are relatively close in the weight space, the algorithm will not converge to minima that are far away. In other words, in order to compare the generalization capabilities of the minima found for a particular network initialized from a particular distribution such that the initializations are close, the Hessian should measure the generalization capability well enough. As an alternative metric to study generalization we also propose the Hessian with respect to the input as introduced in the next section. This metric does not suffer from the scaling symmetries in the weight space of the deep neural network. A way to keep the weight Hessian invariant against these kind of transformations is to consider the weight Hessian multiplied by the weight matrix; note that
\begin{align}\label{eq:scaledhess}
\begin{bmatrix}\alpha (\hat w^{(1)})^T&\alpha^{-1}(\hat w^{(2)})^T\end{bmatrix} &H^w(\mathcal{L}(\alpha W^{(1)},\alpha^{-1}W^{(2)}))\begin{bmatrix}\alpha \hat w^{(1)}\\ \alpha^{-1}\hat w^{(2)}\end{bmatrix} \\
&= \begin{bmatrix} (\hat w^{(1)})^T&(\hat w^{(2)})^T\end{bmatrix}H^w(\mathcal{L}(W^{(1)},W^{(2)}))\begin{bmatrix} \hat w^{(1)}\\  \hat w^{(2)}\end{bmatrix},
\end{align}
where $\hat w^{(1)}$, $\hat w^{(2)}$ denotes the vectorized forms of $W^{(1)}$, $W^{(2)}$, respectively.
The Hessian multiplied by the weight matrix should be resistant against the scaling from \eqref{eq:scaling} and result in a Hessian of similar size as the one of the original, unscaled weights. We study the this metric as a measure for generalization in the numerical experiments in Section \ref{sec6}. 
%From this analysis we see that while for a neural network with trained weights, a large width of the minimum is not necessary for good generalization (i.e. there exist sharp minima that will generalize), a wider minimum is better resistant against noise in the weights, and as we will show in Section \ref{sec42} can also be related to resistance to noise in the inputs. In general we thus expect a wider minimum to have a smaller generalization error than a sharp minimum. 

\subsection{The input Hessian}
Besides the weight Hessian, the input Hessian will also be used as a metric for out-of-sample performance. In this section we discuss the relationship between input noise resistance of the network and the input and weight Hessians. 
\subsubsection{Definition}
The curvature of the loss surface as a function of the weights is given by the Hessian with respect to the weights, as discussed previously. This measures the flatness of particular critical points, and correlates with generalization capabilities. Alternatively, as used in e.g \cite{novak18}, the Jacobian with respect to the input can be used as a measure for generalization. This measures the smoothness of the output function with respect to the input parameters. In \cite{sokolic17} the authors also study this Jacobian as a measure for generalization and propose to explicitly penalize the Frobenius norm of the Jacobian with respect to the training data in the training objective to find minima that generalize well. For this local sensitivity metric one can define the Jacobian $J^x(\hat y)$ with respect to the input $x$ as a vector with elements
\begin{align}\label{eq:jacobi}
j_{i}^x= \partial_{x_i} \hat y(x,w),
\end{align}
where the network is considered to have one output node; it is trivial to extend the definition to multiple outputs. %; in our case since we have a binary categorical output, i.e. the probability of the return going up or down, we have $y_1$ as the probability of an up movement and $y_2$ as the probability of a down movement. 
%While in case of the Hessian we worked with the trace of the matrix as a measure of generalization, here we work with the Frobenius norm and use as a metric of generalization the sensitivity of the output with respect to the input averaged over the data samples,
%\begin{align}\label{eq:jacobinorm}
%\mathbb{E}_{x}\left[||J(x)||_F\right] \approx \frac{1}{N}\sum_{i=1}^N ||J(x^i)||_F.
%\end{align}
An input Jacobian with small elements, would imply that the output function is robust against small changes in the input data, meaning that a better generalization can be achieved. 
 
Besides the Jacobian, the Hessian with respect to the input can also be used. This measures the curvature of the output function or loss function with respect to a varying input. To be consistent with the weight Hessian, we compute the Hessian with respect to the input of the loss function. This measures the curvature of the loss function, and thus the output function, with respect to the inputs, such that a Hessian with small eigenvalues means a smoother output function. Define the elements of the input Hessian $H^x(\mathcal{L})\in\mathbb{R}^{n_0\times n_0}$ of an input $x\in\mathbb{R}^{n_0\times n_0}$ as,
\begin{align}
h_{ij}^x = \partial_{x_i}\partial_{x_j} \mathcal{L}(x,w,y).
\end{align} 
The Jacobian and the Hessian are averaged over the data samples $x^i$, $i=1,...,N$ in the train dataset in order to obtain an average sensitivity metric over the input space. 
%Then the trace of the input Hessian averaged over the data samples will be used as a metric for generalization,
%\begin{align}
%\frac{1}{N}\sum_{i=1}^N Tr(H(x^i)).
%\end{align}
%\begin{remark}[Singularity of the Hessian for the ReLU]
%We remark again that when using the rectified linear unit as an activation function, the second-order derivative of the output with respect to the input is zero. We will use the sofplus, the smooth approximation of the ReLU, in the numerical results. Nevertheless this results in a highly singular Hessian matrix, both with respect to the inputs and the weights. 
%\end{remark}
%\blu{hessian van loss is niet zero, maar geeft dit dan wel iets nuttigs? of is de gewone jacobian genoeg? want het geeft dan alleen product van de eerste afgeleiden. maar in geval van niet relu is alles oke natuurlijk, maar bedenk ff goed wat die hessiaan geeft. }

\subsubsection{The learning rate, batch size and input Hessian}\label{sec43}
In Section \ref{sec312} we showed the effects of the learning rate and batch size on the weight Hessian. Here we show that SGD and its hyperparameters also put restrictions on the input Jacobian and Hessian. Consider SGD where the weight update rule is given by \eqref{eq:wupnoise}. By a first-order Taylor expansion in $x$ for some noise $\tilde\epsilon$ it holds,
%\begin{align}
%\mathcal{L}\left(x+\frac{1}{\sqrt{M}}\epsilon,w,y\right)\approx \mathcal{L}\left(x,w,y\right) + \frac{1}{\sqrt{M}}\epsilon \nabla_x \mathcal{L}\left(x,w,y\right).
%\end{align}
%Therefore, 
\begin{align}
\nabla_{w} \mathcal{L}\left(x+\frac{1}{\sqrt{M}}\tilde \epsilon,w,y\right) \approx \nabla_{w} \mathcal{L}\left(x,w,y\right) + \frac{1}{\sqrt{M}}\nabla_{w} \tilde \epsilon^T \nabla_x \mathcal{L}\left(x,w,y\right)=:\nabla_{w} \mathcal{L}\left(x,w,y\right) + \frac{1}{\sqrt{M}}\epsilon.
\end{align}
In other words, \eqref{eq:wupnoise} can be rewritten as, 
\begin{align}
w_{t+1} \approx w_t - \eta \tilde g, \;\;\; \tilde g :=\mathbb{E}_{(x,y)\sim\mathcal{D}}\left[\nabla_w \mathcal{L}\left(x+\frac{1}{\sqrt{M}}\tilde \epsilon,w,y\right) \right].
\end{align}
Therefore, the noise from the stochastic gradient descent can be related to noise in the input and SGD can be interpreted to minimize a jittered cost function. In turn, by a derivation similar to \cite{reed92}, taking the loss function to be the MSE, one can find,
\begin{align}\label{eq:sgdopt}
\mathcal{L}\left(x+\frac{1}{\sqrt{M}}\tilde \epsilon,w,y\right)\approx \mathcal{L}\left(x,w,y\right) + \frac{||K||_2^2}{M}||\nabla_x \tilde y(w,x)||^2 + \frac{||K||^4_2}{2M} ||H^x(\tilde y(w,x))||_2^2,
\end{align}
where $H^x(\tilde y(w,x))$ is the Hessian with respect to the input $x$ of the neural network output $\tilde y(w,x)$ and the output function in the loss term on the r.h.s. is given by $\tilde y(x,w) = \hat y(x,w) + \frac{||K||}{2M}Tr(H^x(\hat y(x,w)))$. In other words, a relation exists between training with SGD and the minimization of the loss function regularized with the first- and second-order derivatives of the output with respect to the input. Therefore, SGD imposes smoothness assumptions on the output function with respect to the input.

\subsection{Relation between the input and weight Hessian}\label{sec42}
Neural networks are considered to be robust if they are resistant to noise in the input. As mentioned in Section \ref{sec314}, flat minima in weight space are linked to good generalization; furthermore this flatness can be controlled through the learning rate. In this section we study the relation between flatness and the input noise robustness. Previous work \cite{seong18} has also studied this relation, and the authors proposed an optimal learning rate to obtain good generalization. Consider the output of a one-layer neural network,
\begin{align}
\hat y(x,w) = W^{(2)}f\left(W^{(1)}x\right)
\end{align}
where $W^{(2)}\in\mathbb{R}^{1\times n_1}$ (assuming the output $\hat y(x,w)\in\mathbb{R}$) and $W^{(1)}\in\mathbb{R}^{n_0\times n_1}$. Denote by $\hat w^1\in\mathbb{R}^{n_0n_1}$, $\hat w^2\in\mathbb{R}^{n_1}$ the vectorized forms of $W^{(1)}$, $W^{(2)}$. We have,
\begin{align}\label{eq:noisecapt}
\hat y(x +\epsilon,w) = W^{(2)}f(W^{(1)}(x+\epsilon) )=W^{(2)}f((W^{(1)}+\tilde\epsilon)x),\;\;\;\textnormal{ only if   }\;\;\;\; \displaystyle \tilde\epsilon = \frac{ W^{(1)}\epsilon x^T}{||x||_2^2},
\end{align}
where $\epsilon\in\mathbb{R}^{n_0}$ and $\tilde\epsilon\in\mathbb{R}^{n_0\times n_1}$. Let $\hat\epsilon\in\mathbb{R}^{n_0n_1}$ be the vectorized form of $\tilde\epsilon$. Then, 
\begin{align}
\mathcal{L}(x+\epsilon, w,y) = \mathcal{L}(x,w+[\hat\epsilon,0I_{n_1}]^T,y),
\end{align}
so that if the output is resistant to additive noise $\hat\epsilon$ in the first weight matrix $W^{(1)}$ then the output is resistant to additive noise $\epsilon$ in the input. Note that additive noise resistance is just one particular type of input transformation one might be interested in; other types could be e.g. resistance to multiplicative input noise, or more complex transformations of the input space such as translations. 

Consider now the Taylor expansion around the weights,
\begin{align}
\mathcal{L}(x+\epsilon, w,y) =\mathcal{L}(x,w+[\hat\epsilon,0I_{n_1}],y) = \mathcal{L}(x, w,y) + \hat\epsilon^T\nabla_{\hat w^1}\mathcal{L}(x, w,y) + \frac{1}{2}\hat\epsilon^T H^{\hat w^1}(\mathcal{L}(x, w,y))\hat\epsilon.
\end{align}
In order to have a minimum such that $\mathcal{L}(x+\epsilon,w,y) \approx \mathcal{L}(x, w,y)$, i.e. the output function should be resistant to additive noise in the input, the elements of the gradient and Hessian of the first-layer weights need to be small. The smoothness of the output function with respect to the input can be controlled during training in several ways. As we showed in Section \ref{sec312}, the flatness in weight space can be controlled through the learning rate or batch size in the SGD updates. Biasing the optimization algorithm into wider minima in weight space results in smoother functions with lower information complexity in the input space. The width is measured by the weight Hessian, so a lower weight Hessian should give better resistance to additive input noise.  %\blu{hier moet je alleen ff checken of het nu de hessian of hessian times weights is!}

%This is similar to what has been mentioned in previous work e.g. \cite{neyshabur15} \blu{klopt deze citatie??? en klopt wat ik hier claim??} where it was claimed that the Hessian scaled by the weights was better in predicting generalization. 

In the above derivation we thus showed that the stability of the output function with respect to input noise is equivalent to noise resistance in the first layer weights. In order to see the effect of the stability requirement on further layer weights note that the expression in \eqref{eq:noisecapt} can be written as,
\begin{align}
\hat y(x +\epsilon,w) =W^{(2)}f((W^{(1)}+\tilde\epsilon)x)=(W^{(2)}+\tilde\epsilon^2)f((W^{(1)}+\tilde\epsilon^1)x),
\end{align}
for some $\tilde\epsilon^2\in\mathbb{R}^{1\times n_1}$ and such that $\tilde\epsilon^1<\tilde\epsilon$. Let again $\hat\epsilon^1$ , $\hat \epsilon^2$ be the vectorized forms of $\tilde\epsilon^1$, $\tilde\epsilon^2$. Then,
 \begin{align}
\mathcal{L}(x+\epsilon, w,y)= \mathcal{L}(x, w,y) + [(\hat\epsilon^1)^T,(\hat\epsilon^2)^T]\nabla_{w}\mathcal{L}(x, w,y) + \frac{1}{2}[(\hat\epsilon^1)^T,(\hat\epsilon^2)^T] H^{w}(\mathcal{L}(x, w,y))[(\hat\epsilon^1)^T,(\hat\epsilon^2)^T]^T.
\end{align}
From this expression we see that if the Hessian with respect to $W^{(1)}$ is not sufficiently small, for a deeper network the remaining noise can be damped by $W^{(2)}$ if the loss function is sufficiently flat in weight space in the directions of the second-layer weights. This may explain why deep networks can be more robust: the noise that is not fully dampened by sufficient flatness in the first-layer weight directions due to the eigenvectors in directions tangent to the first-layer weights having large eigenvalues  (i.e. sharp increases of the loss function in those directions), can still be dampened in the further layers if the eigenvalues corresponding to the eigenvectors in the directions of the further weights are sufficiently small. On the other hand, adding nodes per layer does not aid in dampening the noise; on the contrary the more nodes per layer the smaller the eigenvalues of the Hessian in all the directions of these layer weights should be. %\blu{beetje wazig verhaal}

%On the other hand, consider a network with a very wide hidden layer $d_2$, versus one with a more narrow one $n_1$, i.e. let $d_1<d_2$. In order to be resistent to additive noise in the input we require the Hessian $\Delta_{w_1}\mathcal{L}(x,w)$ to be small. For the wider network, the Hessian with respect to $d_2$ terms needs to be small. Overfitting is thus easier, since there is a larger probability of some eigenvector having a corresponing large eigenvalue. \blu{check dit ook}

\section{How low can we go?}\label{sec5}
As mentioned before, at least theoretically under particular assumptions, the entropy of a minimum decreases with the loss value. In other words, the lower the loss at a minimum, the larger the trace of the weight Hessian. A similar property can be said to hold for the input Hessian, with functions overfitting on the noise having lower train loss but a higher trace of the input Hessian. The trade-off between optimally representing the data and the smoothness of the function, i.e. the capability to compress the function by dismissing irrelevant input data, is known as the information bottleneck. In the optimal case, a neural network should learn to extract the most informative patterns, with the most compact function possible. 
\subsection{The information bottleneck}
In the work of e.g. \cite{tishby15}, \cite{achille18}, \cite{tishby17} %\cite{alemi17} 
the information bottleneck is studied for neural networks. The information bottleneck is used to extract the most relevant information that the input variable contains about the output variable. Let $I(x;y)$ denote the mutual information where $x$ and $y$ are as usual random variables sampled from some data-generating distribution $\mathcal{D}$,
\begin{align}
I(x;y) = \int_x\int_y p(x,y)\log\left(\frac{p(x,y)}{p(x)p(y)}\right)dxdy,
\end{align} 
where $p(x,y)$ is the joint probability density of $x$ and $y$ and $p(x)$ and $p(y)$ are the marginals of $x$ and $y$ respectively.
The mutual information measures the information that $x$ and $y$ share; i.e. it quantifies the amount of information obtained about one of the random variables by observing the other. Let $\hat y $ be a representation of $x$, such that the distribution of $\hat y $ is fully described by the conditional $p(\hat y |x)$. This representation $\hat y $ is sufficient for $y$ if $I(\hat y ;y)=I(x;y)$, in other words if $\hat y $ contains all the relevenant information $x$ had about $y$. It is minimal if $I(x;\hat y )$ is smallest among the sufficient representations, in other words if the complexity of the represenation is the lowest. The trade-off between the sufficiency and optimality is formulated as the minimization of the information bottleneck Lagrangian,
\begin{align}
L(p(\hat y |x)) = I(\hat y ;x) - \beta I(\hat y ;y),
\end{align}
where $\beta$ operates as the trade-off parameter between the complexity (first term) and the sufficiency (second term). For an overparametrized network to fit a complex dataset (e.g. memorize random noise \cite{zhang16}) it has to pay a price in terms of the information complexity. Bounding the information complexity can thus prevent the overfitting, but the trade-off between the two may depend on the particular dataset used. % \blu{dus alles is alsnog data afhankelijk; dat is mijn punt heir; generieke methoden bestaan niet, alles wat ze laten zien dat werkt op images werkt niet perse bij mij omdat de data structuur zo anders is!}

In the case of neural networks, the representation $\hat y $ is governed by the learned weights $w$, which can be viewed as a random variable depending on the data and the optimization. The mutual information of the weights and the data can be denoted by $I(w;(x,y))$. The flat minima, i.e. the ones with small eigenvalues of the weight Hessian, can be interpreted as having low information. In other words, since the minimum is flat, the weights can be stored at lower precision, requiring fewer bits and having a lower information value. This result is derived more precisely in Proposition 4.3 of \cite{achille18}; here we state their main result. Let $w^*$ denote a local minimum of the cross-entropy loss and $H^w$ is the Hessian at that point. For the optimal choice of the posterior $p(w|x,y)=\epsilon\cdot w^*$, the following bound can be obtained,
\begin{align}
I(w;(x,y)) \leq \frac{1}{2}d\left[\log||w^*||_2^2 + \log ||H^w||_* - d\log(d^2\beta/2)\right],
\end{align}
where $d=\dim(w)$ and $||\cdot||_*$ denotes the nuclear norm. The nuclear norm is given by $||A||_*=Tr(\sqrt{A^* A})=Tr(A)$ for square, real matrices. 
The trace norm of the Hessian is equivalent to the $L_1$-norm of the vector of eigenvalues of the Hessian; therefore minimizing the nuclear norm is the same as reducing the rank of the original matrix (fewer non-zero eigenvalues). %\blu{dan is dit super in tegenstrijd met Chaudhari... waarin die soms hogere rank heeft bij ebtere generelization en soms andersom.. beetje vaag}
This thus states that flat minima have low information. The authors in \cite{achille18} furthermore derive that when decreasing the information in the weights (by some form of regularization), one automatically improves the minimality and thus the invariance of the function. The converse, i.e. low information implies flatness, does not need to hold; in other words, as mentioned in Section \ref{sec314}, there exist minima with good generalization that are not flat. 

\subsection{Dependence on noise}\label{sec52}
%\blu{Hoe kunnen we dit kwantificeren? zegmaar hoe weten we wat noise is? een DFT toepassen? Dynamic time warping? Autocorrelation function? Kunnen we de Hessian wrt input plotten during training? Zouden we daar ook zo een sharp phase transition in zien?}
Consider a time series $y^i$, $i=0,...,N$ with a signal to noise ratio of $(1-\alpha):\alpha$. Suppose the neural network output should be resistant to noise in the signal. \begin{remark}[Robustness in non-i.i.d. setting]
We remark here that in the non-i.i.d. setting, the network can overfit to not just the noise, but also to certain patterns present in one part of the series but not in another. In this paper we mostly focus on robustness to noise, and assume that the non-i.i.d. property comes from the time dependence and the noise which we assume can change in distribution between the train and the test set. Generalization then refers to finding a pattern which is present over time, and the ability to generalize across different noise distributions. A similar setting has been considered for image recognition problems in e.g. \cite{geirhos18}. 
\end{remark}
 In this case the loss function should satisfy the following objective,
\begin{align}
|\mathcal{L}(x+\alpha\epsilon,w,y)-\mathcal{L}(x,w,y)|<\delta.
\end{align}
Relating this to the train and test set, we assume that $x+\alpha\epsilon$ is the test data with a noise component different from that in the train data $x$. In this setting, a small $\delta$ corresponds to a small difference between the error on the train data and the error on the test data, or, in other words, a small generalization error. By a Taylor expansion in the input one obtains,
\begin{align}
\mathcal{L}(x+\alpha\epsilon,w,y)-\mathcal{L}(x,w,y) \approx \alpha\epsilon^T\nabla_x\mathcal{L}(x,w,y) + \frac{1}{2}\alpha^2\epsilon^T H^x(\mathcal{L}(x,w,y))\epsilon.
\end{align}
Then, taking expected values we have,
\begin{align}
\mathbb{E}[\mathcal{L}(x+\alpha\epsilon,w,y)-\mathcal{L}(x,w,y)] \approx \frac{1}{2}\alpha^2Tr\left(H^{x}(\mathcal{L}(x,w,y))\right),
\end{align}
% of is het \begin{align}
%\mathbb{E}[\mathcal{L}(x+\alpha\epsilon)-\mathcal{L}(x,w)] \approx \frac{1}{2}\alpha^2\sum_{i=1}^{n_0}\Delta_{x_i}\mathcal{L}(x,w),
%\end{align}
where we have used the fact that $\epsilon_i\sim\mathcal{N}(0,1)$ i.i.d..
From this it follows that, 
 \begin{align}
Tr\left(H^{x}(\mathcal{L}(x,w,y))\right)=2\frac{\delta}{\alpha^2}.
\end{align}
In other words, the amount of noise in the input the neural network has to be resistant to is inversely proportional to the input (and thus weight) Hessian. This is intuitive in classification problems. Consider an image or a time series one would like to classify. If the output function has to be resistant to $\alpha\epsilon$ noise, i.e. the classification output is invariant to $\alpha\epsilon$ noise in the input, we require the Hessian to be small, so that the shifted input still results in the same output. The higher the noise resistance should be, the smaller the Hessian. The downside is that the requirement for the learned function to posess more resistance against noise can also decrease the performance of the classifier. 
%\subsubsection{Obtaining the signal-to-noise ratio}
%\blu{is dit nodig? of is er een andere link te leggen tussen hessian en noise?}

\subsection{Obtaining better generalizable minima}\label{sec53}
In this section we summarize which hyperparameters can be used to control the trade-off between generalization and complexity (typically the train data fit). These hyperparameters are similar to what is used in an i.i.d. setting, however using the derivations in the previous sections and as we will show in the numerical section in \ref{sec6} these hyperparameters can also control the output in the non-i.i.d. setting. 
\paragraph{Learning rate} In Section \ref{sec312} the relationship between the weight Hessian and the learning rate was discussed. It was shown that using a higher learning rate can result in wider minima if convergence is obtained. In Section \ref{sec43} a the same kind of relationship is derived between the learning rate and the input Hessian. In \cite{seong18} the authors also make a link between a high learning rate and a wide minimum. In particular, they claim that using a high learning rate allows the training algorithm to escape from sharp minima in the weight space so that the optimization algorithm converges to smoother and wider minima that are able to generalize better to unseen data. %In order to avoid the weights exploding, the authors in \cite{seong18} propose to set the learning rate large after the occurence of large gradients, which tend to occur in the beginning of training. 
%The learning rate can be defined to have a Gaussian shape,
%\begin{align}
%\eta_t = \eta_{\max} e^{-\frac{(\tilde t-0.5)^2}{\sigma^2})},\;\;\; \sigma^2=-0.25\frac{1}{\ln(\eta_{\min}/\eta_{\max})},
%\end{align}
%where  $\tilde t:=t/N_{it}$ is the normalized iteration, and $\eta_{\min}$ and $\eta_{\max}$ are the minimum or initial and maximum learning rate, respectively. Using this scheme, the learning rate starts at $\eta_{\min}$, increases to $\eta_{\max}$ in the middle of the training iterations, and then decreases back to $\eta_{\min}$ towards the end of training. 
By starting with a small learning rate the weights do not diverge in the beginning when the gradients tend to be large, but due to the increase in learning rate the weights do not converge to a sharp local minimum either. A similar learning rate schedule was proposed in \cite{smith17sc}, where the authors used a cyclic learning rate with one cycle and a large maximum learning rate. Our derivations in the previous sections thus give a theoretical explanation as to why the learning rate can be used as a control parameter for generalization. In Section \ref{sec6} we study this numerically. In order to avoid the size of the gradient influencing the minima to which the optimization converges (as does happen in the previous works of 
\cite{seong18}), we propose to normalize the gradient by its $L_2$ norm. 

\paragraph{Batch size} The batch size used, similar to the learning rate, determines the size of the  noise of the SGD, as discussed in Sections \ref{sec312} and \ref{sec43}. A smaller batch size results in a larger variance of the noise, which in turn, according to \eqref{eq:sgdopt} results in the smoothing terms, i.e. the input Jacobian and Hessian, having a larger weight in the optimization objective. This makes the trade-off between data fit and function complexity more biased towards obtaining a low function complexity. The batch size can thus determine the amount of smoothness required in the output function learned by the neural network. 

\paragraph{Number of iterations} A small learning rate results in smaller steps taken in the network. In other words, for the same number of training iterations a smaller learning rate may give rise to a minimum with a higher loss value. By a similar argument, a smaller number of training iterations gives rise to a minimum with higher loss. By Theorem \ref{thm2}, at least in theory, the higher-loss minima should also have a higher entropy and thus a better generalization. Therefore, early stopping, or equivalenty training with fewer iterations, terminates the algorithm at a point in the loss surface with higher entropy. Training the network for fewer iterations should avoid overfitting on the noise, and by controlling the train error and the Hessian one can stop training when the sufficient trade-off between fit and smoothness has been obtained. In this way, the number of training iterations can be used to control the smoothness of the obtained solution, i.e. the trade-off in terms of data fit and the information complexity of the learned solution.

%Alternative methods which can result in better generalizable minima are ensembling (a form of smoothing of the output function or the weights), and data augmentation or adding noise to the weights or inputs while training (which can be shown to result in a training objective where as a regularization term the Jacobian or Hessian is added). 

%\subsubsection{Ensembling}
%Consider a general method of averaging over the outputs of $N$ trained neural networks, $y^1(w^1,x),...,y^N(w^N,x)$. Suppose each output satisfies,
%\begin{align}
%y^i(w^i,x)=y(w,x) + \epsilon^i,
%\end{align}
%where $y(w,x)$ is the true underlying signal and $\epsilon^i\sim\mathcal{N}(0,1)$. In other words, each network learns some noisy representation of the true output. The ensemble average,
%\begin{align} 
%\bar y(w,x) = \frac{1}{N}\sum_{i=1}^N y^i(w^i,x),
%\end{align}
%for enough networks $N$ will then average out the noise. Similarly for the Hessian. The larger fluctuations can average out for more networks making the Hessian smaller. 
%\subsubsection{Data augmentation}

\section{Numerical results}\label{sec6}
The neural network we consider, contains $N_{depth}$ hidden layers with $N_{width}$ nodes per layer. The weights are initialized as $\mathcal{N}\left(0,\frac{1}{N_{width}}\right)$ and trained with SGD. The learning rate is set to $\lambda$, with mini-batches of size $N_b$, and $N_{it}$ iterations are used to minimize the mean squared error (MSE). The network is trained to predict the value at time $t+1$ given the time series at times $t-n,t-n+1,...,t$, i.e. $n$ historical datapoints. We use the hyperbolic tangent as the activation function. The results are presented for 20 trained networks starting from different initializations. 
%\paragraph{Norm of the Jacobian and Hessian}
The typical measures of generalization based on the input and weight Hessians are particular norms of the matrices. The trace of the weight Hessian will be used, i.e. 
\begin{align}
Tr(H^w) = \sum_{i=1}^d h^w_{ii}.
\end{align}
 Similarly, the trace of the input Hessian averaged over the data samples will be used as a metric for generalization,
\begin{align}
\mathbb{E}_{x}\left[Tr(H^x)\right] \approx \frac{1}{N}\sum_{i=1}^N Tr\left(H^{x^i}\right).
\end{align}
While in case of the Hessians we work with the trace of the matrices as a measure of generalization, for the input Jacobian we use the Frobenius norm and use as a metric of generalization the sensitivity of the output with respect to the input averaged over the data samples,
\begin{align}\label{eq:jacobinorm}
\mathbb{E}_{x}\left[||J^x||_F\right] \approx \frac{1}{N}\sum_{i=1}^N \big|\big|J^{x^i}\big|\big|_F.
\end{align}

%\paragraph{Difference between the $L_1$ and $L_2$ norm}
%Alternatively, as a measure of the smoothness of the function, the difference between the $L_1$ and $L_2$ norm can be used. \blu{check}

%\paragraph{generalization for piece-wise linear functions}
% Consider the Hessian with respect to the input,
%\begin{align}
%\Delta l(y(w,x)) = l''(y(w,x))\nabla_x y(w,x)\nabla_x y(w,x)^T+l'(y(w,x))\Delta_x (y(w,x)),
%\end{align}
%where $l(y(w,x)) = (y(w,x)-y)^2$, i.e. the per sample squared loss and $l'(y)$ refers to the derivative with respect to the output function $y$. For a piecewise linear function $\Delta_x (y(w,x)=0$. A similar argument holds for the Hessian with respect to the weights where the elements on the diagonal of the weight Hessian will be determined solely by the first-order derivatives. 

\subsection{Artificial data}
In this section we use artifical datasets to gain understanding of the neural network and its generalization capabilities. We show that as expected from the theory, a linear relation exists between the trace of the input and weight Hessians (i.e. the function complexity) and the generalization error. In our results, a higher function complexity means that the network is learning a function which is overfitting on the noise. This is clearly undesirable and results in a worse test set performance. The main task of this section is to show how to recognize when the network starts overfitting, and how to avoid this using the optimization hyperparameters as described in Section \ref{sec53}. 
\subsubsection{Random noise}
\begin{figure}[h!]
  \centering
  \includegraphics[width=0.35\textwidth]{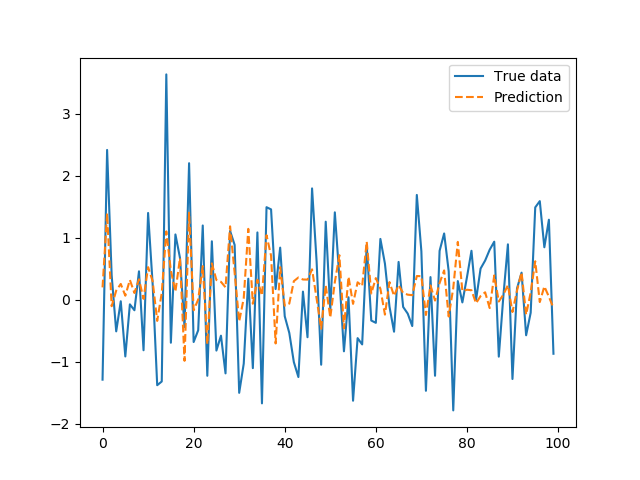}
  \includegraphics[width=0.35\textwidth]{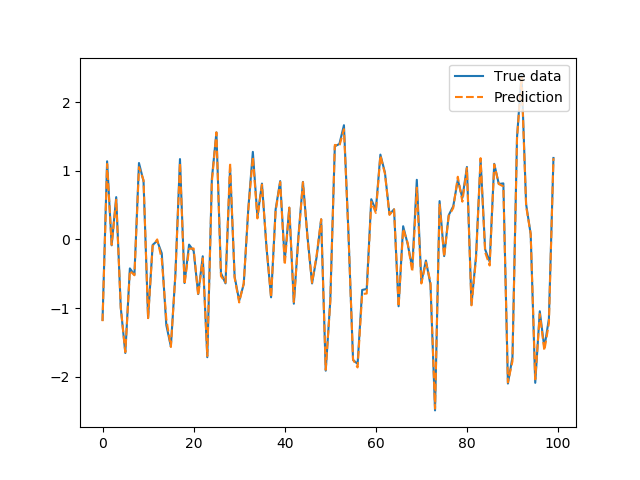}\\
    \caption{The initial fit (L) and the output function (on the train data) learned (R) by a neural network of size $N_{depth}=1$ and $N_{width}=500$ with $\lambda=0.05$ on random noise data. An overparametrized network trained with SGD can fit to random noise. This effect is undesired and ideally, a network will not converge on noise.}\label{fig01}
\end{figure}

\begin{figure}[h!]
  \centering
  \includegraphics[width=0.49\textwidth]{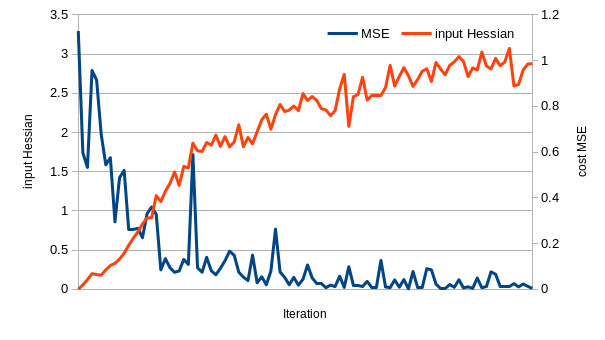}
  \includegraphics[width=0.49\textwidth]{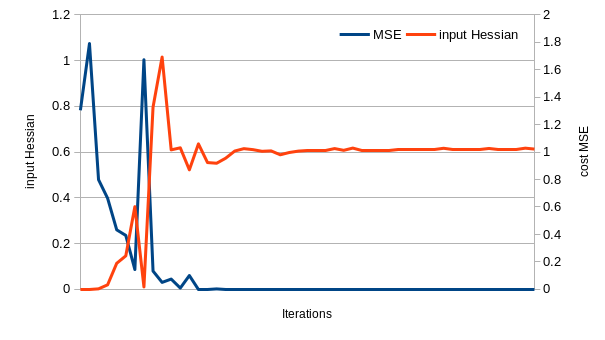}\\
  \includegraphics[width=0.49\textwidth]{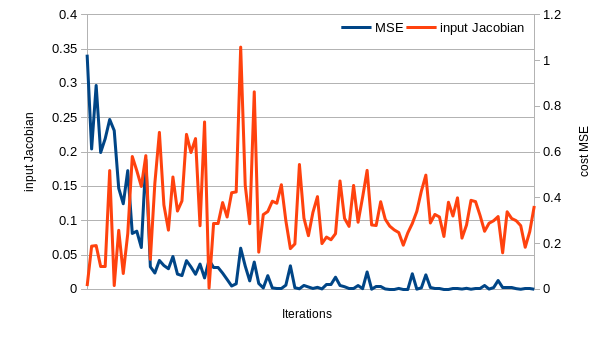}
  \includegraphics[width=0.49\textwidth]{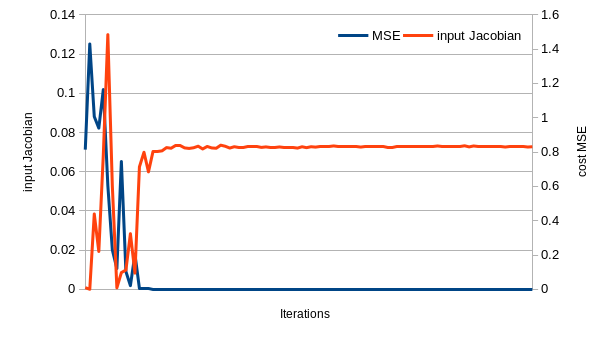}\\
    \caption{The convergence of the network of size $N_{depth}=1$ (L) and $N_{depth}=10$ (R) and $N_{width}=500$ with $\lambda=0.05$ for the input Hessian (T) and the input Jacobian (B) on random noise data; the loss decreases with iterations, but the input norms of the Jacobian and Hessian increase, as the output function increases in complexity. The input Jacobian and Hessian can give indication for when a network starts overfitting on the noise. This can then be avoided by making a trade-off between complexity and train error.}\label{fig03}
\end{figure}

\begin{figure}[h!]
  \centering
  \includegraphics[width=0.49\textwidth]{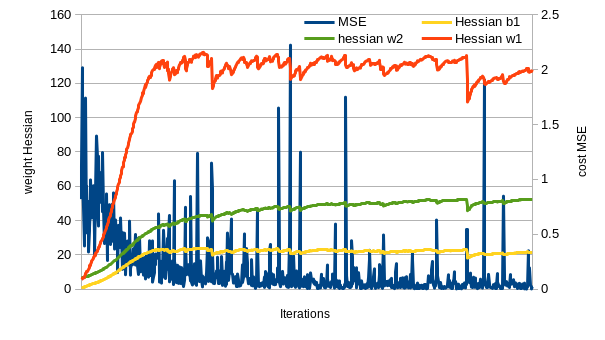}
  \includegraphics[width=0.49\textwidth]{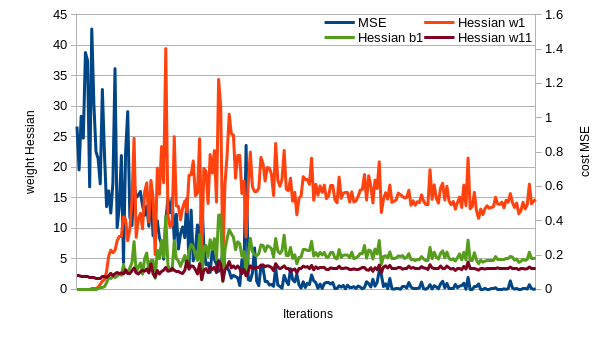}\\
    \caption{The convergence of the network of size $N_{depth}=1$ (L) and $N_{depth}=10$ (R) and $N_{width}=500$ with $\lambda=0.05$ in terms of the loss function and the trace of the weight Hessians per layer on random noise data; the loss decreases with iterations, but the weight Hessians increase, showing that in order to obtain low loss the function complexity has to significantly increase. The weight Hessian norm can be used in order to bound the function complexity in order to avoid convergence on noise.}\label{fig04}
\end{figure}

We simulate 100 datapoints from an $\mathcal{N}(0,1)$ distribution. An overparametrized neural network with the number of parameters larger than the sample size will be able to perfectly fit this random noise, see Figure \ref{fig01}. However, in order to obtain a low loss, the network will have to significantly increase its function complexity, as can be measured by the norms of the weight and input Jacobians and Hessians. This is shown in Figures \ref{fig03}-\ref{fig04}: the traces of the input and weight Hessians significantly increase to obtain a small loss value. After some number of iterations, if the MSE has converged to some value, the Hessian remains approximately constant, with small fluctuations due to the stochasticity of the optimization algorithm. When the network starts to learn the high-frequency components, here the noise, the norms of the input and weight Hessians increase significantly. In order to thus avoid convergence on noise one has to keep these norms small. 

%Note that the Hessians of a deeper network (both w.r.t. input and weight) are smaller than those of the shallow network. In terms of generalizibility this does not mean anything since the average MSE over 50 networks for $N_{depth}=1$ is 1.46 and the MSE for $N_{depth}=10$ is 1.78. From this we can conclude that the trace of the Hessian alone is not enough of an indicator; while we do not expect to be able to obtain a linear relation between the Hessian and test set error due to the data being fully random, the input Hessian should be comparable across different network structures \blu{is dit zo? waarom is die nou kleiner dan? je vermenigvuldigt in meer lagen vaker f' en f'' en de weights, dus als deze klein zijn krijg je dit?}. 

%\begin{figure}[h!]
%  \centering
%  \includegraphics[width=0.49\textwidth]{randomnoise_1_500_100samples_test.png}
%  \includegraphics[width=0.49\textwidth]{randomnoise_10_500_100samples_test.png}\\
%    \caption{The output function (on the test data) learned by a neural network of size $N_{depth}=1$ (R) and $N_depth=10$ (R) with $N_{width}=500$ and $\lambda=0.05$. The average MSE over 50 networks for $N_{depth}=1$ is 1.46 and the MSE for $N_{depth}=10$ is 1.78.}\label{fig01}
%\end{figure}

\subsubsection{Noisy sine function}
Consider now the function $y_i = \sin(0.1t_i)+c\epsilon_i$, with $t_i\in\{0,1,...,100\}$ and $\epsilon_i\sim\mathcal{N}(0,1)$. The network input consists of $(y_{i-4},...,y_{i})$ and is trained to forecast $y_{i+1}$. Note that this time series is clearly not stationary, but contains seasonality which is a common feature of time series such as weather measurements. 

\begin{figure}[h!]
  \centering
    \includegraphics[width=0.49\textwidth]{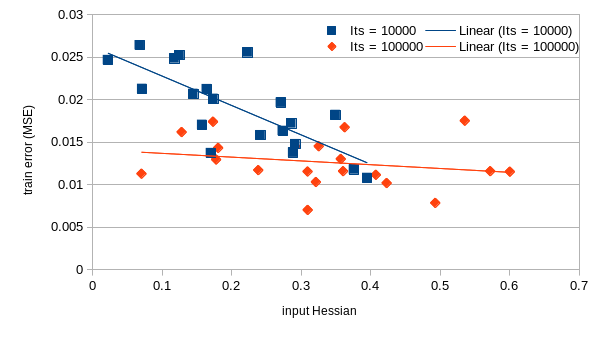}
  \includegraphics[width=0.49\textwidth]{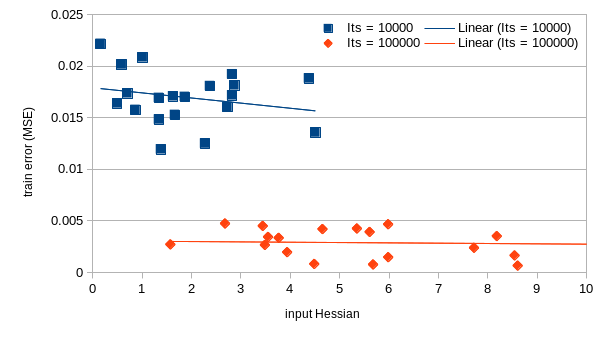}
    \includegraphics[width=0.49\textwidth]{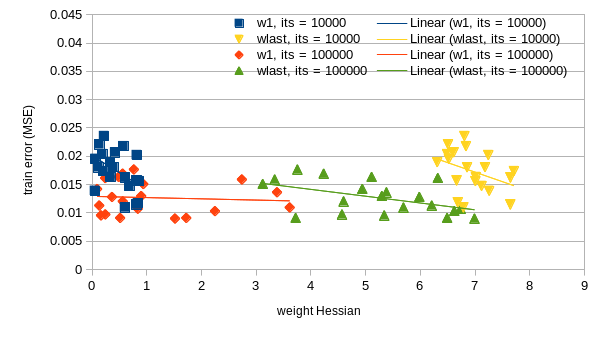}
    \includegraphics[width=0.49\textwidth]{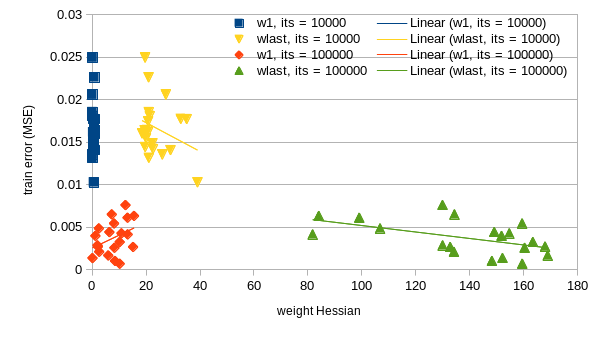}
    \caption{The train error and trace norm of the input Hessian (T) and the weight Hessian (B) for the noisy sine with $c=0.1$ for 20 trained networks. The neural network has 1 (L) and 10 (R) layers with 500 nodes per layer and is trained for a different number of iterations (10000 and 100000). Training longer results in a solution of higher complexity as measured by the norms of the input and weight Hessians. This solution has a smaller train error but will likely result in worse generalization.}\label{fig05a}
\end{figure}

\begin{figure}[h!]
  \centering
    \includegraphics[width=0.49\textwidth]{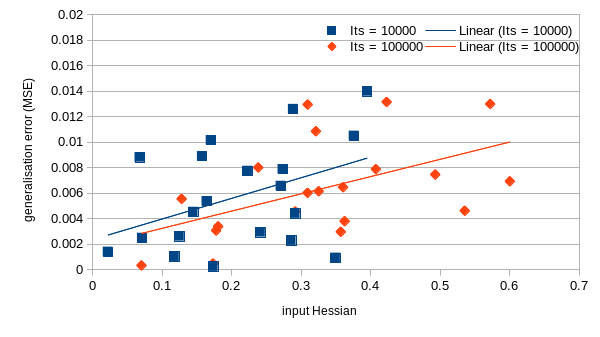}
  \includegraphics[width=0.49\textwidth]{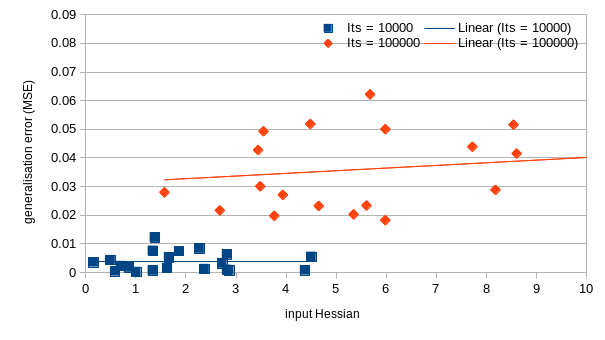}
  \includegraphics[width=0.49\textwidth]{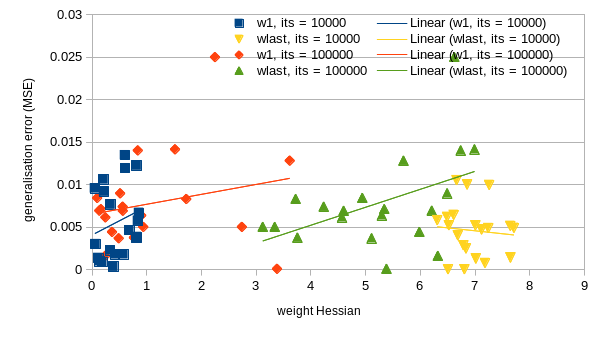}
  \includegraphics[width=0.49\textwidth]{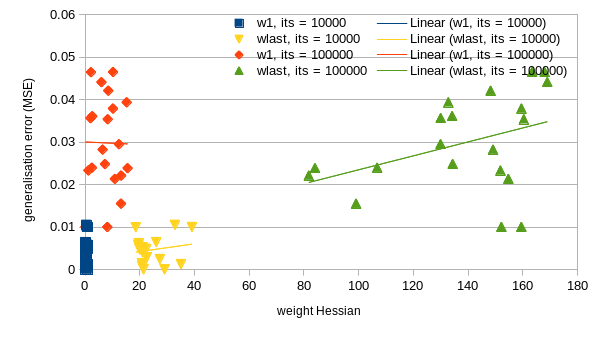}
    \caption{The generalization error and trace norm of the input Hessian (T) and the weight Hessian (B) for the noisy sine with $c=0.1$ for 20 trained networks. The neural network has 1 (L) and 10 (R) layers with 500 nodes per layer and is trained for a different number of iterations (10000 and 100000). A smaller trace of the Hessian results in a lower generalization error, and training longer increases the complexity of the learned solution.}\label{fig05b}
\end{figure}

\paragraph{Generalization and the number of iterations} Figure \ref{fig05a} and \ref{fig05b} shows the trace norm of the input and weight Hessians plotted against the train error and the generalization error, respectively. There exists a linear relation between the trace of the input and weight Hessians and the generalization error: a smaller trace norm results in lower generalization error. This effect is slightly less significant for the deeper network. Furthermore, training longer results in a solution of higher complexity, which in this case is undesired since a higher complexity means that the function is overfitting on the noise. This is in accordance with the theoretical result on the entropy and the loss in Theorem \ref{thm2}, where it was claimed that the lower the train loss, the sharper the minimum, or the larger the output function complexity is. Training longer allows to access lower points in the loss surface with a lower train error as seen in Figure \ref{fig05a}. These lower points have a lower entropy, which results in a higher generalization error as seen from Figure \ref{fig05b}.  %Furthermore, for deep neural networks the Theorem states that this trade-off is less severe. From the right-hand side of Figure \ref{fig05a} and \ref{fig05b} this indeed holds: lower loss minima do not differ much in terms of minima width compared to higher loss minima, validating again the theoretical results from \cite{becker18}. 
%We remark that with more layers, the first layer weights do not appear to always be an indicator of train and generalization error: a higher trace of the Hessian with respect to the first layer weights results in a higher train error and lower generalization error, the opposite of the expected relationship. In other words, this could mean that, as observed in Section \ref{sec52}, the sensitivity of the output function with respect to the first layer weights is of less significance to generalization, i.e. even if the loss surface in the directions of the first weights shows sharp increases, this does not mean that a large generalization error will be attained. 

%\begin{figure}[h!]
%  \centering
%  \begin{minipage}{0.35\textwidth}
%    \includegraphics[width=\linewidth]{plot_train_1_sine01_lowHessian.png} 
%\footnotesize
%Test error: 0.013, trace of input Hessian: 0.17
%    \end{minipage}
%    \begin{minipage}{0.35\textwidth}
%    \includegraphics[width=\linewidth]{plot_train_1_sine01_highHessian.png} 
%\footnotesize
%  Test error: 0.016, trace of input Hessian: 0.25
%    \end{minipage} 
%    
%        \caption{The learned functions for the sine with $c=0.1$ and a network of one layer, with 1000 iterations (L) and 10000 iterations (R). Training longer results in a more complex solution with larger generalization error; early stopping can result in a smoother solution with better generalization.}\label{fig05d}
%\end{figure}

\begin{figure} [h!]
 \centering  
    \begin{minipage}{0.35\textwidth}
    \includegraphics[width=\linewidth]{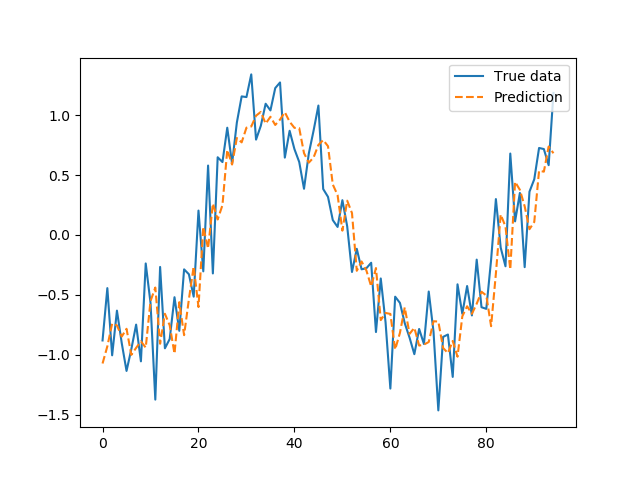}
\footnotesize
Test error: 0.13, trace of input Hessian: 0.013
    \end{minipage}\hspace{0.5cm}
    \begin{minipage}{0.35\textwidth}
    \includegraphics[width=\linewidth]{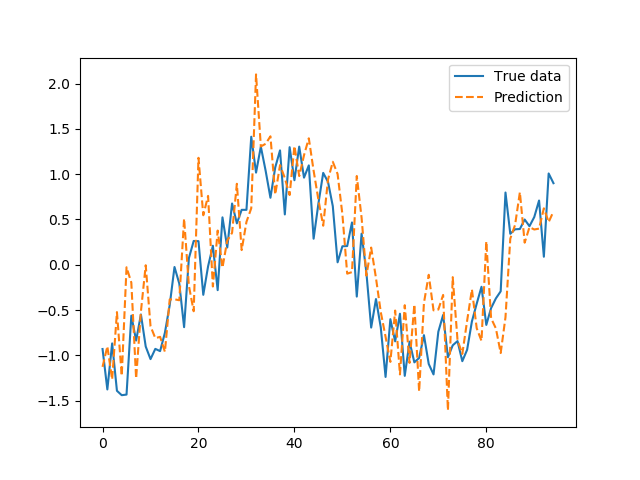}
\footnotesize
  Test error: 0.30, trace of input Hessian: 1.41
  \end{minipage}
     \begin{minipage}{0.35\textwidth}
    \includegraphics[width=\linewidth]{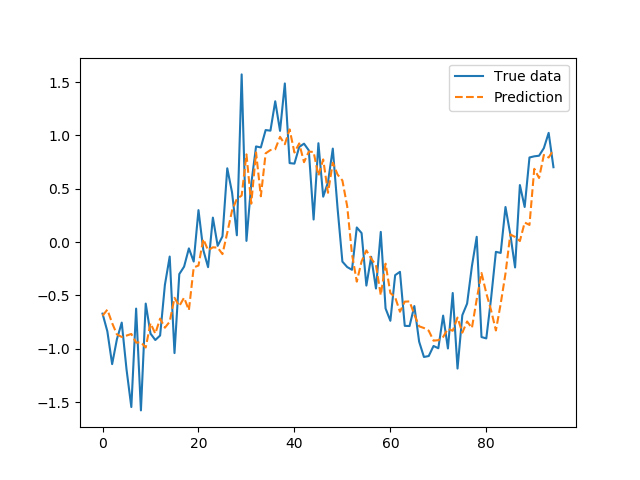}
\footnotesize
Test error: 0.14, trace of input Hessian: 0.004
    \end{minipage} \hspace{0.5cm}   
    \begin{minipage}{0.35\textwidth}
    \includegraphics[width=\linewidth]{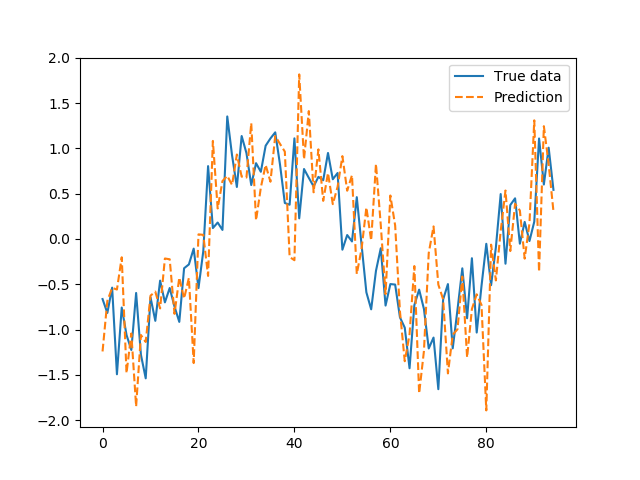}
\footnotesize
  Test error: 0.43, trace of input Hessian: 29.1
    \end{minipage}

    \caption{The functions learned by the network on the sine data with $c=0.3$ with 1000 (L) and 10000 (R) iterations with a network of one layer (T) and 10 layers (B). With more noise the network is prone to overfitting, especially in the deep network. A smaller number of training iterations results in a lower Hessian which clearly results in a smoother function with lower test error.}\label{fig05e}
\end{figure}

\paragraph{Increasing the noise amplitude}
Consider now the sine function with the noise coefficient given by $c=0.3$. In Figure \ref{fig05e} we observe that in order to obtain a generalization error in the high noise case ($c=0.3$) similar to that in the low noise case ($c=0.1$, Figure \ref{fig05b}) the Hessian should be much smaller. This corresponds with the theoretical analysis in Section \ref{sec52}, where it was observed that with more noise in the signal one requires a lower Hessian in order to obtain a similar generalization error. Finding a low complexity solution on noisy data can be difficult due to the possibility of overfitting, since no explicit smoothness contraints are imposed. Deep networks are even more prone to overfitting due to the higher number of parameters and the sharper gradient descent directions. Thus, in order to obtain sufficiently smooth solutions for deep networks one needs to adapt the training method or cost function accordingly. For the training method, as seen in Figure \ref{fig05e} taking fewer steps -- or equivalently (not shown in the plots but discussed in Section \ref{sec53}) using smaller learning rates-- results in smaller generalization error. 

%\begin{figure}[H]
%  \centering
%    \includegraphics[width=0.47\textwidth]{sine_noise03_1_500_20_hessian_input_its.png}
%  \includegraphics[width=0.47\textwidth]{sine_noise01_10_500_20_hessian_input_its.png}
%    \caption{The generalization error and trace norm of input Hessian for the noisy sine with $c=0.3$ for 20 trained networks. The neural network has 1 (L) and 10 (R) layers with 500 nodes per layer and is trained for a different number of iterations (10000 and 100000). A smaller trace of the Hessian results in a lower generalization error, and training longer increases the complexity of the learned solution.}\label{fig05c}
%\end{figure}

\begin{figure}[h!]
  \centering
    \includegraphics[width=0.47\textwidth]{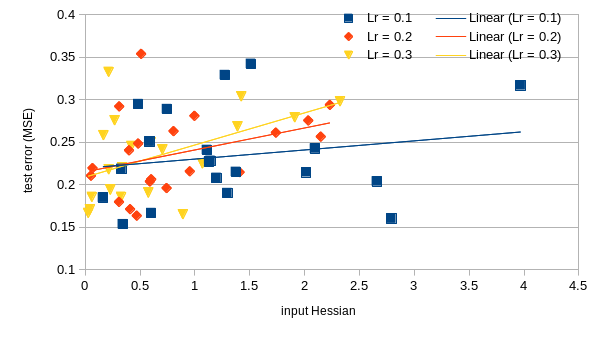}
  \includegraphics[width=0.47\textwidth]{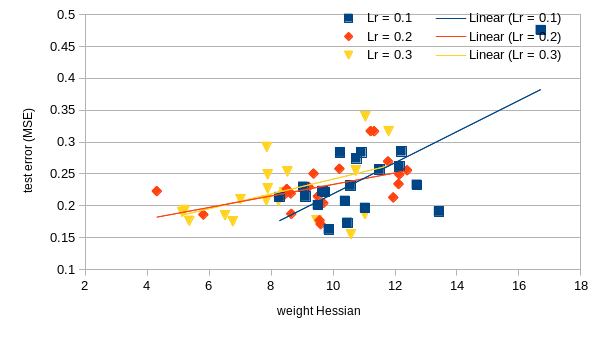}
   \includegraphics[width=0.47\textwidth]{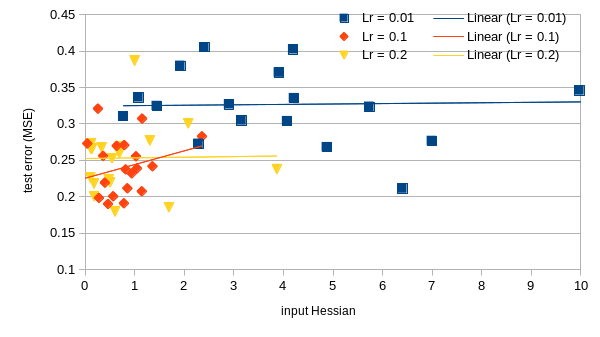}
     \includegraphics[width=0.47\textwidth]{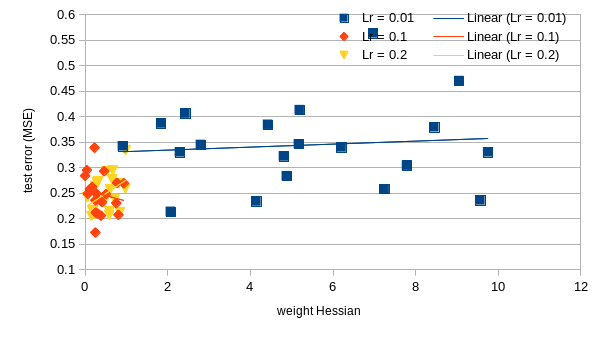}
    \caption{The test error and the input (L) and weight (R) Hessians for the  noisy sine function with $c=0.3$ for different learning rates. The neural network has 1 (T) and 10 (B) layers with 100 nodes per layer. On average, training with a larger learning rate results in wider minima, i.e. smaller input and weight Hessians, however significant overlap between the minima found with different learning rates nevertheless exists.}\label{fig06a}
\end{figure}

\paragraph{Generalization and the learning rate}
Here we study the effects of the learning rate on generalization. Figure \ref{fig06a} shows the test error plotted against the input and weight Hessians obtained by training the neural network with different learning rates. Using a larger learning rate results in wider minima while a smaller learning rate tends to converge to sharper minima, however a significant amount of outliers are found in both cases. We used batch gradient descent and scaled the gradient by its $L_2$ norm -- i.e. the gradient in the SGD updates is given by $g/||g||_2$ -- in order to avoid the gradient size influencing the minima width. %When the learning rate is large but the gradient is small one might still converge to sharp minima. Similarly, when the learning rate is large and the gradient is also large the weights can diverge. We furthermore use full batch gradient descent so that the noise from the stochastic gradients does not influence the results. 
We remark that the relation between a larger learning rate and wider minima seems to be more clear in the deep neural network where the usage of the higher learning rates results in more minima clustered at lower values of the trace. While the Hessian is correlated with the test error, i.e. a small test error means a smaller Hessian, the test set performance is not significantly better using the higher learning rates. Even though convergence has been obtained, the network weights have converged to a minimum that underfits the data and therefore the output function can have a worse test set performance. 

\begin{figure}[h!]
  \centering
    \includegraphics[width=0.47\textwidth]{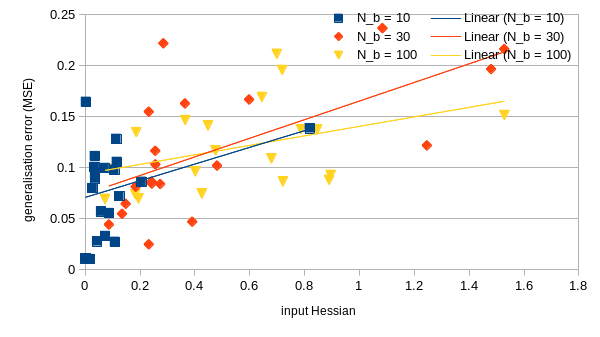}
  \includegraphics[width=0.47\textwidth]{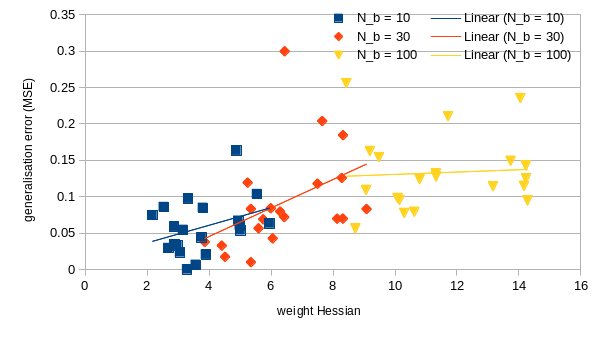}
   \includegraphics[width=0.47\textwidth]{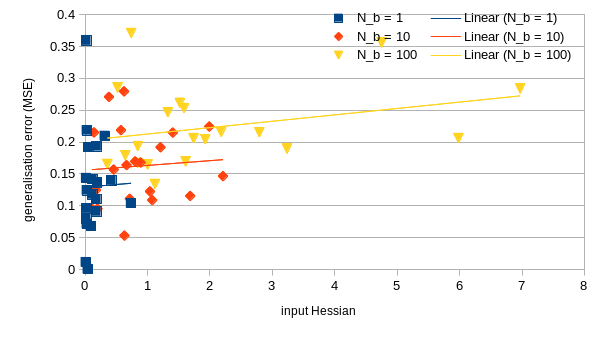}
     \includegraphics[width=0.47\textwidth]{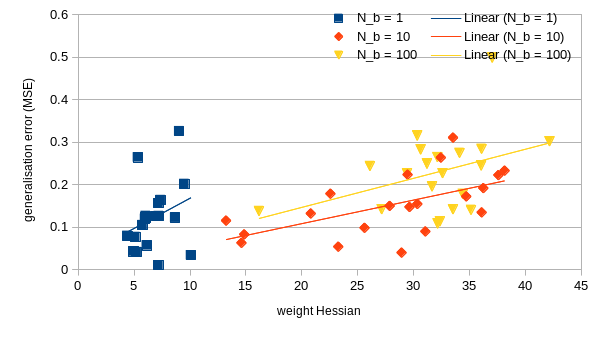}
    \caption{The generalization error and the trace of the input (L) and weight (R) Hessians for the noisy sine function with $c=0.3$ for different batch sizes. The neural network has 1 (T) and 10 (B) layers with 100 nodes per layer. Training with a smaller batch size results in a minimum with smaller values of its input and weight Hessian, which means a smoother output function and a wider minimum in weight space. Smaller batch sizes thus result in functions which can generalize better.}\label{fig07a}
\end{figure}

\paragraph{Generalization and the batch size} In Figure \ref{fig07a} we plot the generalization error and the traces of the input and weight Hessians for different batch sizes. Using a smaller batch size causes the network to converge to minima with lower input and weight Hessians, which in turn correspond to minima with lower generalization error. As expected from the theory in Section \ref{sec53} batch size has a significant influence on the smoothness of the output function with respect to input and weights, and can be used as a control for the trade-off between train and generalization error. Out of the three controls considered: number of iterations, learning rate and batch size, the number of iterations and the batch size appear to be the most effective ones for controlling the trade-off. 
  %In Section \ref{sec53} it was mentioned that the batch size determines the amount of noise in the SGD algorithm, and is related to the weight given to the smoothness requirements in the SGD optimization objective.  

\begin{figure}[h!]
\centering
\includegraphics[width=0.47\textwidth]{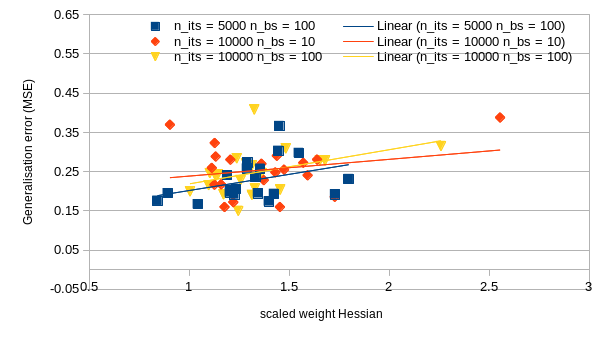}
\includegraphics[width=0.47\textwidth]{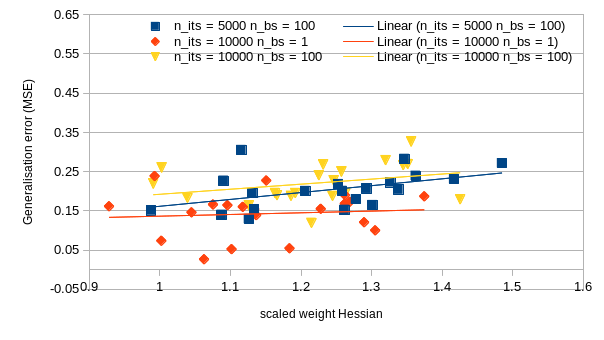}
\caption{The generalization error with respect to the weight Hessian multiplied by the weight vector as defined in \eqref{eq:scaledhess} for a network with 1 (L) and 10 (R) layers for the noisy sine function with $c=0.3$. A linear trend is observed with a smaller Hessian giving a lower generalization error. The scaled Hessian as a measure for generalization seems to be as accurate as the unscaled Hessian, showing that the scaling sensitivity is not a significant problem for the minima found with SGD.}\label{fig07b}
\end{figure}

\paragraph{The scaled Hessian as a metric for generalization}
Here we use the Hessian multiplied by the weights as defined in \eqref{eq:scaledhess} as a measure for generalization. While good results were obtained with the original weight Hessian, it is of interest to see if the amount of outliers (here minima with different Hessians but similar generalization erros) decreases when using the scaled weight Hessian. In Figure \ref{fig07b} we see that similar to the unscaled weight Hessian a clear linear dependence exists between the size of the Hessian and the generalization error. This dependence does not seem to be more significant as compared to the unscaled Hessian, showing that as expected from Section \ref{sec23}, the scaling sensitivity of the weight Hessian is not a significant problem for the minima found by SGD, and the weight Hessian is a valid metric for measuring generalization capability, despite its scaling sensitivity.

\subsection{Real-world data}
In this section we study generalizability for several real-world time series forecasting. We show that the norm of the input and weight Hessian is a good metric for measuring the capability of a network to generalize, and that similar to the artifical dataset, the hyperparameters defined in Section \ref{sec53} are very effective for controlling the trade-off between smoothness -- as measured by the Hessians -- and data fit -- as measured by the train loss.
\begin{remark}[Network architecture]
In the coming examples we consider a network architecture with $N_{depth}=2$ and  $N_{width}=100$. Similar results, with regard to the effects of the controls and the ability of the metrics to measure generalization, hold for other architectures as long as the network is overparametrized. The hyperparameters would have to be tuned accordingly to the network size. For example, the more parameters a network has, the easier it is to overfit so that even fewer iterations might be needed to avoid the overfitting. Furthermore, as has been mentioned in e.g. \cite{park19} the wider the neural network, the more noise it can handle, which in turn results in better generalization. 
\end{remark}

\subsubsection{Index data}
Financial data is highly non-linear, non-stationary and has a very low signal-to-noise ratio \cite{cont01}. Overfitting on the training data and not being able to generalize well to unseen data is therefore a challenge. We will use a network of size $N_{depth}=2$, $N_{width}=100$. The input data will consist of $n=5$ historical daily absolute returns of the S\&P500 index, $n=5$ historical daily absolute returns of the CBOE 10 year interest rate, and $n=5$ historical daily absolute returns of the volatility index (VIX), so that the total input into the neural network will consist of 15 nodes. The train period consists of data from 2017-01-03 until  2018-02-02, and the test data from 2018-02-03 until 2018-08-13. Given the value of the time series $S_t$ the returns are computed as $r_t = S_{t} - S_{t-1}$. The returns are then normalized using the mean and variance. The network output will consist of the prediction for the next day return $r_{t+1}$ of the S\&P500 index. In Table \ref{tab01} we present the MSE and hit rate (computed as the number of up or down movements predicted correctly) for different hyperparameters averaged over 20 sampled networks. A larger trace of the input or weight Hessian appears to correspond to a worse performance; similarly, training longer results in overfitting. A smaller batch size corresponds to a smaller weight Hessian in the final layer, but it does not seem to result in better performance due to e.g. underfitting the signal. Financial returns are highly noisy and non-linear and distinguishing the signal in the data from noise remains challenging. Nevertheless we showed that the techniques presented in the paper can be used to bias the algorithm into minima that have more (additive) noise resistance. 

\begin{table}[H]
\centering
\begin{tabular}{ l|l||l|l|l|l|l }
$N_{it}$ & $N_{b}$ & MSE & hit rate & $Tr(H^x)$ & $Tr(H^{W^{(1)}})$& $Tr(H^{W^{(3)}})$\\\hline\hline
  10000&100&2.80&0.49&0.73&5.60&45.16\\
  5000&100 & 2.65 & 0.513 & 0.73 & 4.86 & 45.48\\
  10000&300&2.76&0.500&0.83&4.81&47.26\\
  5000&300&2.71&0.505&0.74&4.50&46.60
\end{tabular} 
\caption{The MSE and hit rate for different batch sizes and iteration numbers. A higher trace of the input or weight Hessian results in a lower hit rate and a higher MSE. Training longer results in a higher error and using a smaller batch size results in a smaller weight Hessian in the final layer, but does not seem to correspond to a better performance due to e.g. underfitting.}\label{tab01}
\end{table}

\subsubsection{Temperature}
\begin{figure}[h!]
  \centering
  \begin{minipage}{0.35\textwidth}
    \includegraphics[width=\linewidth]{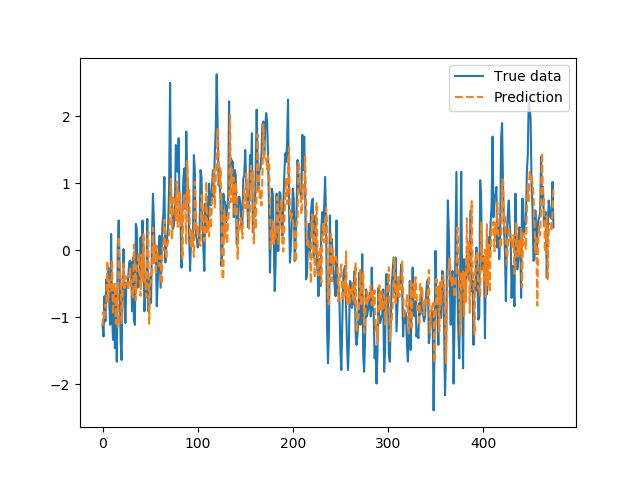}
\footnotesize
Test error: 0.38 
    \end{minipage}
    \begin{minipage}{0.35\textwidth}
    \includegraphics[width=\linewidth]{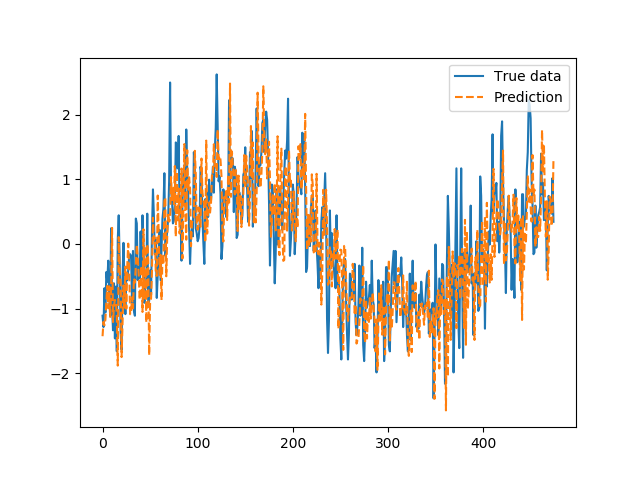}
\footnotesize
Test error: 0.50
    \end{minipage}   
  
    \caption{The temperature forecast for a network trained for 5000 iterations (L) and 10000 iterations (R). There is a clear seasonality in the dataset, however the network that was trained longer overfits on the noise in the obervations and therefore generalizes worse. }\label{fig09a}
\end{figure}

In this section we train a network for predicting the daily minimum temperature in Melbourne, Australia. The dataset contains observations over the period of 1988-01-01 until 1990-12-31. We will use a network of size $N_{depth}=2$, $N_{width}=100$. The input data will consist of $n=20$ historical daily obervations of the temperature. The results for the MSE for different training hyperparameters are presented in Table \ref{tab02} and Figure \ref{fig09a}. As expected, training with fewer iterations and using smaller batch sizes results in a smoother output function with respect to the input and causes the training algorithm to converge to wider minima. The temperature data has a clear seasonal pattern but the daily obervations vary due to noise; using smaller batch sizes or training shorter has a regularising effect on the output function, so that the network does not overfit on the noise in the data, but continues to follow the main trend.

\begin{table}[h!]
\centering
\begin{tabular}{ l|l||l|l|l|l }
$N_{it}$ & $N_{b}$ & MSE & $Tr(H^x)$ & $Tr(H^{W^{(1)}})$& $Tr(H^{W^{(3)}})$\\\hline\hline
10000&10&0.44&0.078&6.05&34.9\\
  5000&10 & 0.36&0.023&5.85&27.5\\  
  10000&100&0.49&0.11&36.7&40.2\\
  5000&100 & 0.36&0.030&38.4&31.9\\
  10000&200&0.50&0.10&42.6&40.9\\
  5000&200&0.37&0.032&56.7&31.7
\end{tabular} 
\caption{The MSE for different batch sizes and iteration numbers. A higher trace of the input or weight Hessian corresponds to a worse test set MSE due to overfitting on the noise. As usual, training longer and using larger batch sizes resuls in more overfitting.}\label{tab02}
\end{table} 

%\subsection{Other data}

\section{Conclusion}
In this work we studied generalization capabilities of neural networks trained for the purpose of time series forecasting. We showed that there is a correspondence between good generalization capability and small input and weight Hessians of the loss function at the minima found after training. A small input or weight Hessian corresponds to the smoothness of the trained function, or, in other words, the resistance of the output function to noise in the input or weights, respectively. The challenge lies in finding the optimal tradeoff between fit of the data and smoothness of the learned function, so as to avoid overfitting on the noise and underfitting on the signal of interest. We showed how to use the learning rate, the batch size and the number of iterations used in the training algorithm to bias the network into minima that posess a certain structure. Other aspects that may influence generalization capabilities are the kind of activation function used: while not reported we noticed that the network is prone to overfitting when using the piecewise linear ReLU compared to the hyperbolic tangent or the sigmoid function. Furthermore, the network size itself also matters: deep networks, due to the larger amount of parameters, will more easily overfit on the noise, obtaining a low training error but a bad out-of-sample performance. 

While this work provided some insight into obtaining good generalization for time series forecasting, forecasting remains a challenging task due to the non-linear and non-stationary distribution of the data. The typical assumption in statistical learning theory of having i.i.d. samples from some data-generating distribution does not hold in time series: there is a dependence between the obervations through time and the underlying distribution may change due to unobserved variables so that the train and test data might not be identically distributed. A relaxation that has become standard to deal with the independence assumption is to assume that the observations are drawn from a stationary mixing distribution (see e.g. \cite{agarwal13}, \cite{mcdonald17}). The authors of e.g. \cite{kuznetsov18}, \cite{kuznetsov17}, \cite{kuznetsov14} provide bounds that also hold for non-stationary time series. A related issue is that of generalization in neural networks across different noise distributions. As has been mentioned in the work of \cite{geirhos18}, neural networks have trouble generalizing when the noise distribution in the data they were trained on differs from the noise distribution in the test dataset. Understanding and solving this issue will prove valuable in time series forecasting, where the distribution of the noise in the observations could change over time. Obtaining theoretical results on e.g. the link between the generalization error and the Hessian, and understanding how to make machine learning algorithms work in order to generalize in a non-i.i.d. setting is still a relevant and active topic of research which we aim to address in future work.

\bibliographystyle{siam}
\bibliography{biblio}

\end{document}